\documentclass[conference]{IEEEtran}
\IEEEoverridecommandlockouts

\usepackage{cite}           
\usepackage{amsmath,amssymb,amsfonts}

\usepackage{graphicx}
\usepackage{subfigure}      
\usepackage{tikz}           
\usetikzlibrary{patterns}   

\usepackage{booktabs}       
\usepackage{multirow}
\usepackage{tabularx}
\usepackage{array}
\usepackage{diagbox}
\usepackage{colortbl}       
\usepackage{rotating}       

\usepackage{textcomp}
\usepackage{xspace}
\usepackage{pifont}          
\usepackage{stfloats}       
\usepackage{microtype}      
\usepackage{url}            
\usepackage[bookmarks=false]{hyperref}
\hypersetup{colorlinks=true,linkcolor=black,citecolor=black,urlcolor=black} 
\usepackage{balance}        
\usepackage[most]{tcolorbox} 

\def\BibTeX{{\rm B\kern-.05em{\sc i\kern-.025em b}\kern-.08em
    T\kern-.1667em\lower.7ex\hbox{E}\kern-.125emX}}

\newif\ifblindreview
\blindreviewfalse   

\newcommand{\algname}{GhostShell\xspace}
\newcommand{\rbtname}{CoCo\xspace}

\newcommand{\eg}{\emph{e.g.}\xspace}         
\newcommand{\ie}{\emph{i.e.}\xspace}         
\newcommand{\etc}{\emph{etc.}\xspace}        
\newcommand{\vs}{\emph{vs.}\xspace}          

\newcommand{\cmark}{\textcolor{green!70!black}{\ding{51}}}  
\newcommand{\xmark}{\textcolor{red!80!black}{\ding{55}}}     

\definecolor{c-body}{RGB}{224,102,102}     
\definecolor{c-head}{RGB}{106,168,79}      
\definecolor{c-bgm}{RGB}{241,194,50}       
\definecolor{c-emotion}{RGB}{109,158,235}  
\newcommand{\cbody}[1]{\textcolor{c-body}{\textbf{#1}}}
\newcommand{\chead}[1]{\textcolor{c-head}{\textbf{#1}}}
\newcommand{\cbgm}[1]{\textcolor{c-bgm}{\textbf{#1}}}
\newcommand{\cemotion}[1]{\textcolor{c-emotion}{\textbf{#1}}}

\begin{document}

\title{GhostShell: Streaming LLM Function Calls\\for Concurrent Embodied Programming}

\ifblindreview
  \author{Anonymous Authors}
\else
  \author{Jian Gong$^{*1}$, Youwei Huang$^{*1\dagger}$, Bo Yuan$^{*1}$, Ming Zhu$^{*1}$,\\
    Zhou Liao$^{1}$, Jianhang Liang$^{1}$, Juncheng Zhan$^{1}$, Jinke Wang$^{1}$, Hang Shu$^{1}$, Zihao Xu$^{1}$,\\
    Mingyue Xiong$^{1}$, Yanjun Ye$^{1}$, Yufan Zu$^{1}$, Yang Zhou$^{1}$, Yihan Ding$^{1}$,\\
    Xuannian Chen$^{1}$, Xingyu Lu$^{1}$, Runjie Ban$^{1}$, and Bingchao Huang$^{1}$%
    \thanks{$^{1}$All authors are with LeapWatt, Shenzhen, China.}%
    \thanks{$^{\dagger}$Corresponding Author. \texttt{\href{mailto:yw@leapwatt.com}{yw@leapwatt.com}}}%
    \thanks{$^{*}$Equal Contribution.}%
  }
\fi

\maketitle

\begin{abstract}
  We present \algname, a novel approach that leverages Large Language Models~(LLMs) for streaming and concurrent behavioral programming in embodied systems.
In contrast to predefined behavioral structures and plan-then-execute paradigms, \algname enables reasoning-while-acting by incrementally invoking functions during LLM streaming generation.
We define function tokens as an XML-based function-call representation that \algname parses from the LLM generation stream and maps to callable functions.
A multi-channel scheduling algorithm further orchestrates these calls with intra-channel synchronous and inter-channel asynchronous dispatch, coordinating sequential-parallel behavior execution across multiple robotic components.
We evaluate \algname on our robotic prototype \rbtname across 33 real-world tasks with LLMs from nine providers.
On 30 grounded Human-Robot Interaction~(HRI) tasks, our approach achieves the highest Directed Structured Behavior Correctness~(DSBC) score of 0.83 with Claude-Sonnet-4, while on three long-horizon multimodal tasks, GPT-4.1 attains a top human evaluation score of 7.0/10.
Compared to native LLM function calling, our function token schema achieves a 15/15 task completion rate versus 6/15, particularly in coordinating concurrent linguistic and physical actions.
Supplementary materials including videos are available at \url{https://coco-robot.github.io/GhostShell}.
\end{abstract}


\section{Introduction}\label{sec:introduction}
Embodied intelligence is widely recognized as a critical step toward artificial general intelligence~(AGI)~\cite{goertzel2014artificial,morris2023levels,bubeck2023sparks}, requiring robots not only to understand and respond to fluent natural language, but also to seamlessly coordinate and manipulate their physical bodies to accomplish real-world tasks.
This demands the ability to process multimodal inputs and perceive both external environments and internal states for context-aware decision-making~\cite{sheridan2016human,bartneck2024human,wang2024lami,koubaa2025next}.
Despite progress across Behavior Trees~(BTs)~\cite{colledanchise2018behavior,lykov2024llm,izzo2024btgenbot}, Task and Motion Planning~(TAMP)~\cite{garrett2021integrated}, Reinforcement Learning~(RL)~\cite{schulman2017proximal,nachum2018data}, Imitation Learning~(IL)~\cite{hussein2017imitation,brohan2022rt,chi2025diffusion}, and Vision-Language-Action~(VLA) models~\cite{zitkovich2023rt,kim2024openvla,sapkota2025vision}, current robots still fall short of human-like behaviors due to three fundamental limitations:
(1) most agents decouple planning and execution, preventing human-like reasoning-while-acting through streaming inference;
(2) coordinating sequential and parallel execution across multiple components (\eg, head, limbs, speech) remains difficult;
(3) existing methods lack generalizability, typically requiring retraining or redesign to accommodate new hardware or tasks.

Large Language Models~(LLMs), especially Large Multimodal Models~(LMMs)~\cite{achiam2023gpt,team2023gemini}, offer foundational capabilities that can address these limitations.
Having been trained on vast corpora of human knowledge, LLMs can comprehend real-world inputs and generate coherent outputs in both natural and programming languages in a streaming manner~\cite{openai2024streaming}.
Prior work has explored LLMs as high-level planners for robotic tasks~\cite{ahn2022can,huang2022inner,liang2023code,liu2024enhancing}, demonstrating strong generalizability and zero-shot task transfer.
However, these approaches still rely on a plan-then-execute paradigm and do not address the coordination of concurrent multi-channel robot behaviors.
This raises an intriguing possibility:
\textit{Can we leverage LLM streaming generation to achieve human-like robotic interaction?}

We present \textbf{\algname}, an LLM-driven approach for streaming and concurrent behavioral programming in embodied systems.
We define \textit{function tokens} as an XML-based function-call representation where the tag name maps to a function name and attributes encode typed parameters.
\algname processes LLM output through a four-stage pipeline:
(1) incrementally parsing \textit{function tokens} and textual tokens from the generation stream;
(2) resolving each element to a callable function and deriving its execution lifecycle from the XML tree structure;
(3) dispatching function calls across component channels with intra-channel synchronous and inter-channel asynchronous scheduling;
and (4) executing each call on platform-specific backends, producing sequential and parallel behavior across multiple components, with observations returned to the LLM for ReAct-style reasoning.

\begin{figure}[ht]
    \centering
    \includegraphics[width=0.85\columnwidth]{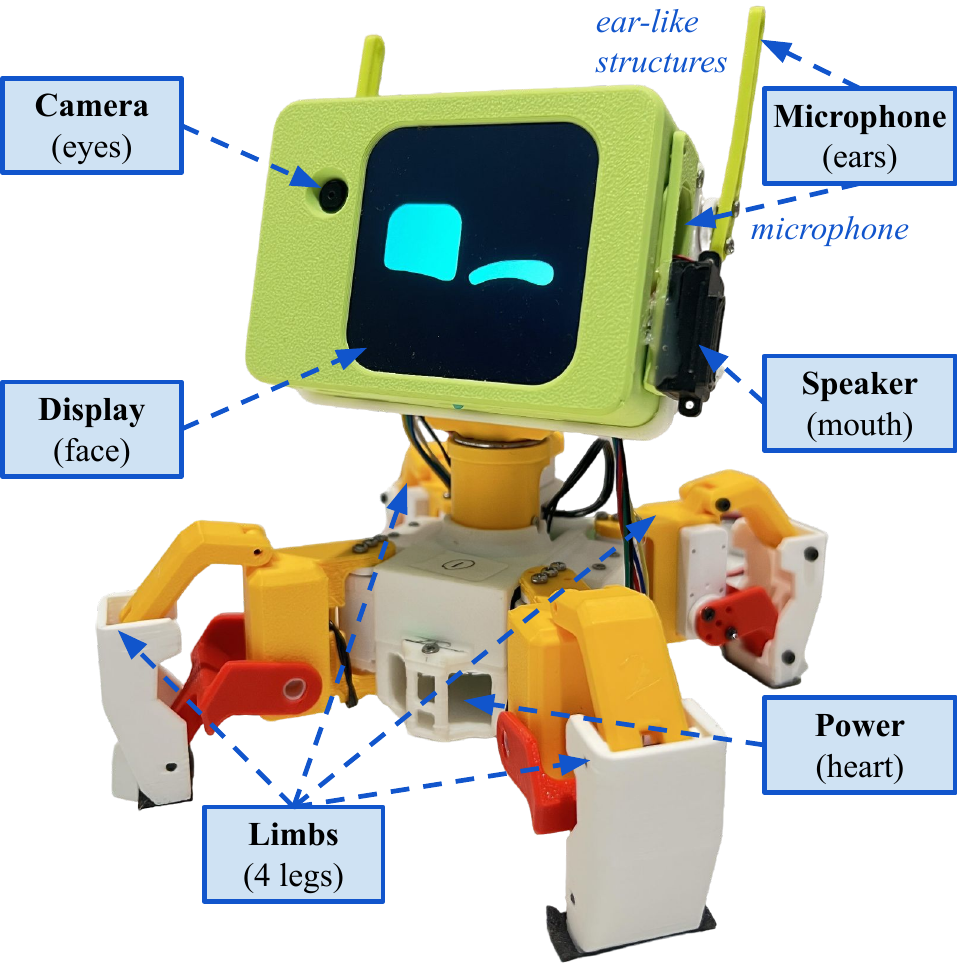}
    \caption{The \rbtname robot: a quadrupedal robot with 12 servos, equipped with a screen (face), speaker (mouth), microphone (ears), and camera (eyes).}
    \label{fig:robot}
\end{figure}

We evaluate \algname on a quadruped spider-like robot \rbtname, shown in Fig.~\ref{fig:robot}, through three experiment groups:
(1) 30 grounded Human-Robot Interaction~(HRI) tasks spanning five behavioral modes, evaluated across 21 LLMs from nine providers using the Directed Structured Behavior Correctness~(DSBC) metric;
(2) three long-horizon multimodal tasks requiring multi-turn reasoning with visual feedback in a ReAct-style loop, assessed via human evaluation on a 10-point scale;
and (3) an ablation study comparing our \textit{function token} design against LLM native function calling.
Among the 30 HRI tasks, \textbf{Claude-Sonnet-4} achieves the highest overall DSBC score of \textbf{0.83}.
On the three long-horizon tasks, \textbf{GPT-4.1} attains the best human evaluation score of \textbf{7.0}/10.
Furthermore, our \textit{function token} approach achieves a perfect 15/15 task completion rate, whereas native function calls succeed on only 6/15 tasks, demonstrating the structural expressiveness of our streaming \textit{function token} design for hierarchical behavioral composition.

Our main contributions are as follows:
\begin{enumerate}
    \item We propose \textit{function tokens}, an XML-based function-call representation that enables on-the-fly action invocation during LLM streaming generation, achieving reasoning-while-acting in embodied systems.
    \item We design a multi-channel scheduling algorithm that performs intra-channel synchronous and inter-channel asynchronous dispatch, enabling sequential-parallel execution across multiple robotic components.
    \item We deploy \algname on a physical quadruped robot \rbtname and conduct extensive real-world evaluation across 33 tasks with 21 LLMs from nine providers, demonstrating practical viability across five behavioral modes of increasing complexity.
\end{enumerate}

\section{Related Work}\label{sec:related}
\subsection{Structured Control and Learning-Based Methods}
Early robot control approaches rely on structured representations and learning-based methods.
BTs~\cite{colledanchise2018behavior} offer modular, interpretable task structures but require manual topology design, limiting adaptability.
TAMP~\cite{garrett2021integrated} enables symbolic-geometric planning but demands exhaustive precondition specification and domain-specific modeling.
RL~\cite{schulman2017proximal,nachum2018data} acquires policies through environment interaction but suffers from poor sample efficiency and limited cross-task generalization.
IL~\cite{hussein2017imitation,brohan2022rt,chi2025diffusion} learns from expert demonstrations but remains constrained by distribution shift and high data collection cost.
VLA~\cite{zitkovich2023rt,kim2024openvla} unifies perception and action generation in a single model but requires large-scale embodiment-specific training data.

These methods serve as task-specific motor primitives that lack open-ended reasoning and cross-platform generalization.
Rather than replacing them, \algname acts as a deliberative cognitive layer atop these primitives, abstracting them into callable functions and coordinating their execution across heterogeneous platforms without retraining.

\subsection{LLMs for Embodied Planning and Control}
LLMs have introduced a new paradigm for robot planning and control~\cite{zeng2023large,wang2025large}.
Early work established zero-shot task planning~\cite{huang2022language}, affordance-grounded LLM execution~\cite{ahn2022can}, and closed-loop planning via environmental language feedback~\cite{huang2022inner}.
Subsequent efforts extended this to manipulation via vision-language grounding~\cite{driess2023palme,gao2024physically}.
To improve generalizability, many approaches encode robot behaviors as LLM-generated executable code~\cite{singh2022progprompt,liang2023code,vemprala2024chatgpt} or behavior trees~\cite{lykov2024llm,izzo2024btgenbot}.
Recent work further addresses uncertainty alignment~\cite{ren2023robots}, world-state tracking~\cite{yoneda2024statler}, interactive planning under partial observability~\cite{sun2024interactive}, bimanual orchestration~\cite{chu2024large}, multi-agent coordination~\cite{yuan2025remac}, long-horizon execution~\cite{mon2025embodied}, multimodal HRI~\cite{wang2024lami}, and LLM-ROS integration~\cite{mower2024ros,koubaa2025next}.

However, most of these works target robotic arms or simple mobile platforms and adopt a plan-then-execute paradigm, requiring complete LLM generation before any action, which inhibits real-time responsiveness.
TypeFly~\cite{chen2023typefly} and TimelyLLM~\cite{ling2024timelyllm} reduce streaming latency for drone control, but remain limited to single-channel sequential execution.
In contrast, \algname parses \textit{function tokens} directly from the LLM generation stream to invoke actions on-the-fly, achieving reasoning-while-acting.
Its multi-channel scheduling algorithm enables simultaneous coordination of heterogeneous components (\eg, speech, locomotion, and facial expressions) within a single embodied agent through formal synchronous-asynchronous dispatch, going beyond single-channel streaming~\cite{chen2023typefly,ling2024timelyllm} and prior multi-modal coordination~\cite{wang2024lami} that lack streaming-level scheduling.

\section{Ghost in the Shell}\label{sec:algorithm}
This section presents \textbf{\algname (Ghost in the Shell)}, where the ``ghost'' is the LLM that reasons and generates structured outputs, and the ``shell'' is the runtime that parses, maps, schedules, and executes these outputs across the robot's physical components.
We describe the overall architecture, the XML \textit{function token} schema, interface-level code abstraction as prompt, and the four-stage shell pipeline.

\subsection{Architecture Overview}\label{sec:arch}

\begin{figure*}[t]
	\centering
	\includegraphics[width=\textwidth]{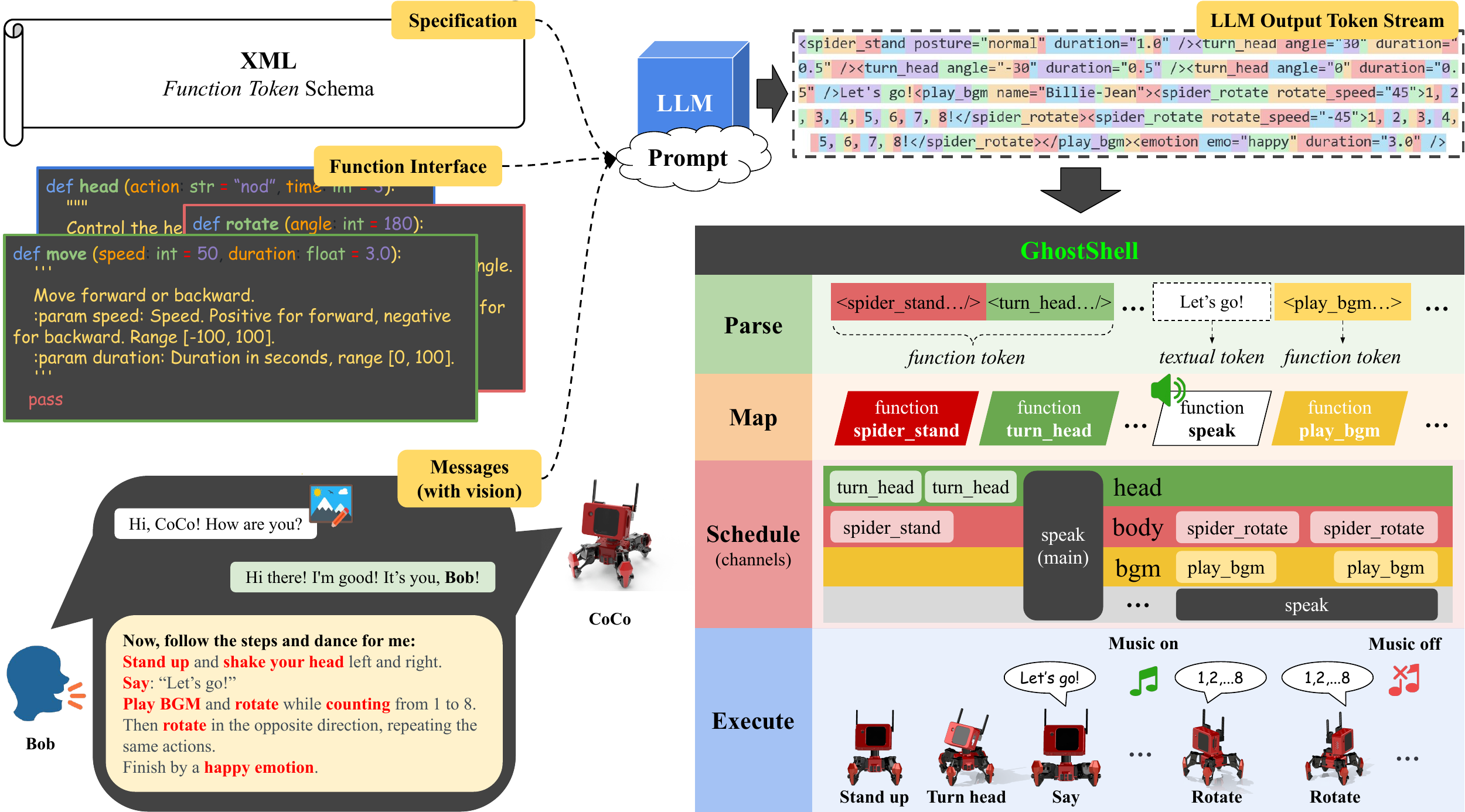}
	\caption{The architecture of \algname embedded in the \rbtname robot:
		Multi-modal prompts are processed by the LLM to produce an output token stream.
		\algname incrementally parses \textit{function tokens} and textual tokens,
		maps them to corresponding callable interfaces in \rbtname,
		and performs multi-channel scheduling across components such as the \chead{head}, \cbody{body}, and \cbgm{audio}.
		The bottom right shows examples of \rbtname's real-time execution results.}
	\label{fig:framework}
\end{figure*}

We embed \algname into a quadruped robot prototype \rbtname and illustrate the overall architecture in Fig.~\ref{fig:framework}.
The system operates as a four-stage pipeline: (1)~parsing the LLM output stream to extract \textit{function tokens} and textual tokens, (2)~mapping both token types to the robot's callable interfaces with parameter binding, (3)~scheduling the resulting function calls across multiple component channels (\eg, \chead{head}, \cbody{body}, \cbgm{audio}) for concurrent execution, and (4)~executing the scheduled calls on the physical hardware in real time.
Processing begins as soon as the first output token arrives, enabling on-the-fly behavioral execution during LLM generation.

We demonstrate the pipeline with a representative interaction.
\rbtname, equipped with visual perception, recognizes user Bob upon greeting.
Bob then issues the following compound instruction:

\begin{tcolorbox}[colback={rgb,255:red,253;green,246;blue,227}, colframe={rgb,255:red,181;green,166;blue,127}, boxrule=0.5pt, arc=2pt, left=6pt, right=6pt, top=4pt, bottom=4pt]
	\textit{\textbf{\textcolor{black}{Now, follow the steps and dance for me:}}
			\cbody{Stand up} and \chead{shake your head} left and right.
			Say: ``Let's go''.
			\cbgm{Play BGM} and \cbody{rotate} while counting from 1 to 8.
			Then \cbody{rotate} in the opposite direction, repeating the same actions.
			Finish by a \cemotion{happy emotion}.}
\end{tcolorbox}

This instruction, together with the \textit{function token} specification and \rbtname's function interfaces, constitutes the system prompt sent to the LLM.
As shown in the bottom-right of Fig.~\ref{fig:framework}, the resulting execution involves \rbtname \cbody{standing up} on the \cbody{body} channel in parallel with \chead{turning its head} on the \chead{head} channel (inter-channel parallelism with intra-channel sequencing), followed by \cbody{rotating} while \cbgm{playing BGM} and counting from 1 to 8, with both actions terminating upon counting completion (lifecycle-scoped termination).
This example exercises three of the five behavioral modes evaluated in Section~\ref{sec:experiment} (sequential, parallel, and lifecycle-scoped termination), illustrating \algname's ability to coordinate complex nested behaviors during generation from a single natural language instruction.

\subsection{XML Function Token Schema Specification}\label{sec:xml-schema}

\begin{figure*}
    \centering
    \includegraphics[width=\textwidth]{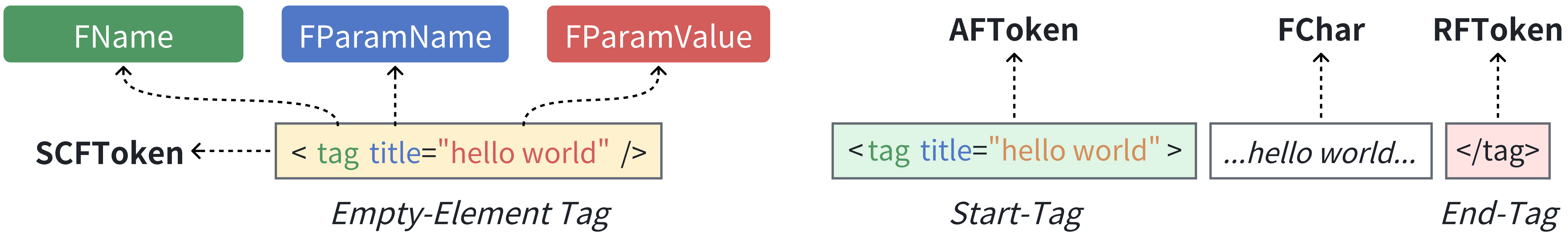}
    \caption{Terminology definitions of XML \textit{function tokens} according to the XML 1.0 specification.}
    \label{fig:xml-func-token}
\end{figure*}

To the best of our knowledge, there is currently no standardized specification for \textit{function tokens} used in streaming function calling.
In this work, we introduce an XML-based \textit{function token} schema for representing function mapping and calling in LLM streaming outputs.

Distinct from standard LLM output tokens produced by subword tokenizers~\cite{sennrich2015neural} and from special reasoning tokens used for chain-of-thought processing~\cite{wei2022chain,guo2025deepseek,zhang2025reasoning}, a \textit{function token} in our formulation represents a higher-level semantic unit corresponding to a function call.
A single function token may span multiple LLM output tokens and is parsed incrementally during streaming generation (refer to Section~\ref{sec:parser}).

Following the XML 1.0 specification recommended by W3C~\cite{bray2000extensible}, we define the terminology and structure of XML \textit{function tokens} used in this work.
As illustrated in Fig.~\ref{fig:xml-func-token}, the schema defines three primary token types: \textit{Activation Function Token} (AFToken, corresponding to the XML start-tag), \textit{Reset Function Token} (RFToken, end-tag), and \textit{Self-Contained Function Token} (SCFToken, empty-element tag), along with \textit{Function Character Token} (FChar) for literal character data and \textit{Function Reference Token} (FRef) for reserved character encoding.
In our schema, the tag name (FName) of each XML element corresponds to the function name, while the parameter names (FParamName) and parameter values (FParamValue) map to the function's input arguments.
The complete formal definitions are provided in Appendix~\ref{appendix:xml-schema}.

\subsection{Code Abstraction as Prompt}\label{sec:code-as-prompt-main}

\algname exposes only \textit{abstract function interfaces} to the LLM via the system prompt.
Each interface declares the function name, typed parameter signatures with value constraints, and a semantic description (\eg, docstring), while all implementation bodies are omitted.
Two equivalent formats are supported: a code-based format using function stubs with typed parameters and docstrings, and an XML-based format using the \textit{function token} schema defined in Section~\ref{sec:xml-schema}.
Either format provides the LLM with sufficient information to discover available capabilities and generate correct function calls without access to any underlying implementation.

At runtime, abstract interfaces are bound to concrete implementations via dependency injection, and functional modules can be dynamically loaded or unloaded with their interfaces automatically reflected in the prompt.
Since the abstraction operates strictly at the interface level, \algname can orchestrate functions across different programming languages, heterogeneous platforms, and diverse backends, including external models, system APIs, third-party programs, and other \algname agents.
Further details and examples are provided in Appendix~\ref{sec:code-as-prompt}.

\subsection{Shell}

The shell enables streaming parsing, dynamic mapping, multi-channel scheduling, and observable execution of content incrementally generated by LLMs, beginning as soon as the first token arrives.
We detail each stage below.

\subsubsection{Parser}\label{sec:parser}

The parser employs a SAX (Simple API for XML) streaming event-driven algorithm to extract \textit{function tokens} from the LLM output stream according to the schema defined in Section~\ref{sec:xml-schema}.
Incomplete \textit{function tokens} are buffered, while each completed token is immediately forwarded to the mapper.
Additional implementation details of the \algname parser are provided in Appendix~\ref{appendix:parser}.

\subsubsection{Mapper}

\begin{figure}[ht]
	\centering
	\includegraphics[width=\columnwidth]{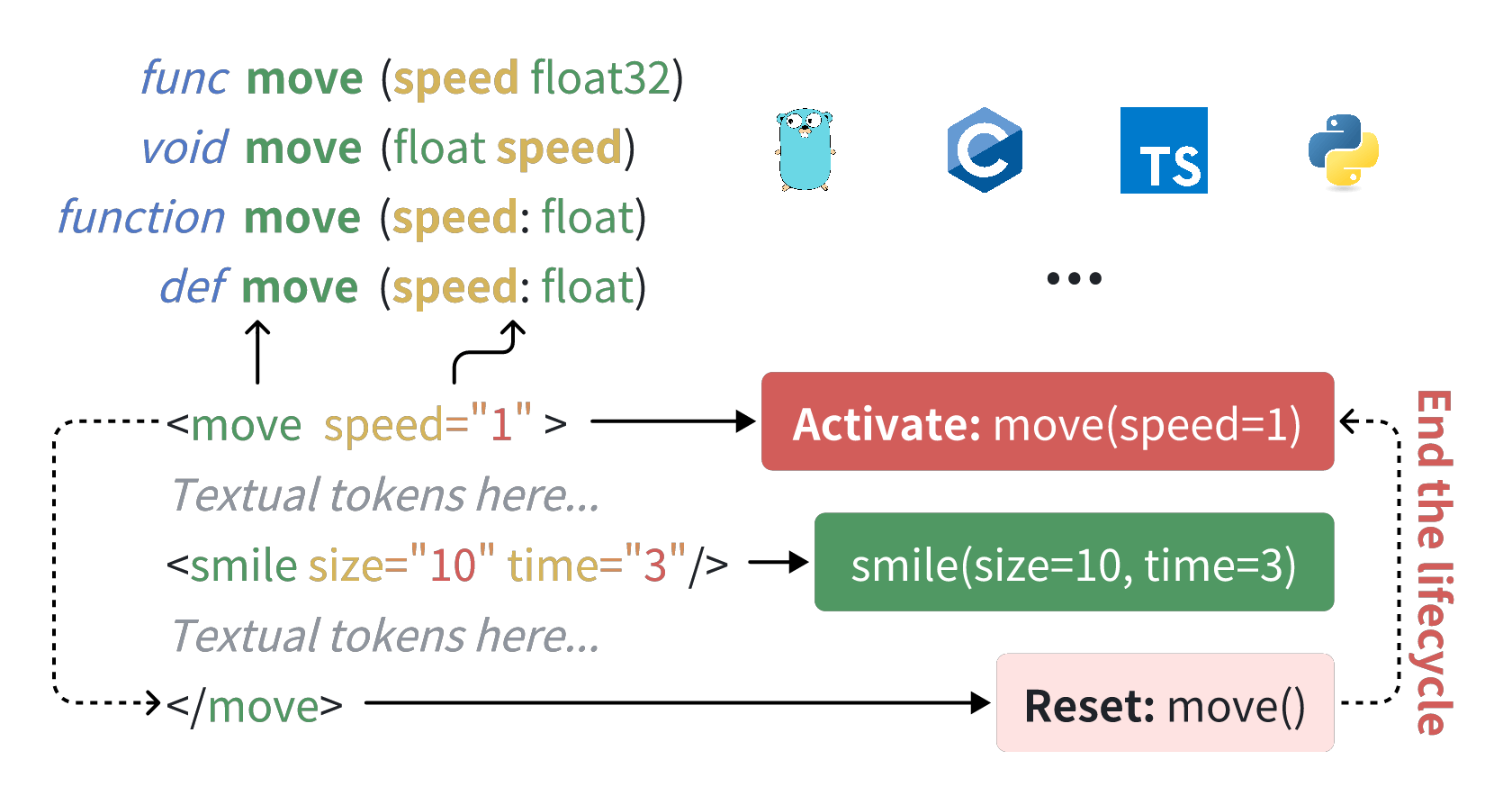}
	\caption{Mapping XML \textit{function tokens} to callable functions with attribute-to-parameter conversion, and the corresponding lifecycle transitions.}
	\label{fig:xml-map-func}
\end{figure}

As illustrated in Fig.~\ref{fig:xml-map-func}, the mapper performs two mappings: it maps each XML element to a callable function with typed parameters, and defines their execution lifecycles through XML's tree structure.

\textbf{Element-to-Function Mapping.}
For each parsed element $E$, the mapper resolves its FName to a function $F$ and converts the FParamName$=$FParamValue attribute pairs into a typed parameter vector $\vec{\theta}$: $\mathcal{M}(E) = (F, \vec{\theta})$,
where $\vec{\theta} = [\theta_1, \ldots, \theta_n]$ are converted to the types required by $F$.

\textbf{Element-to-Lifecycle Mapping.}
The mapper associates each element $E$ with a tree-structured lifecycle $\mathcal{L}(E)$ governing activation and termination.
For a \textit{non-empty element} (bounded by AFToken and RFToken):
\begin{equation}
	F:
	\Sigma_0
	\xrightarrow[\text{activate}]{\text{AFToken}}
	\Sigma_{\text{active}}
	\xrightarrow{\{\mathcal{L}(E_i)\}_{i=1}^k}
	\Sigma_{\text{active}}
	\xrightarrow[\text{reset}]{\text{RFToken}}
	\Sigma_0
\end{equation}
where $\{E_i\}_{i=1}^k$ ($k \geq 0$) are nested child elements whose lifecycles are recursively resolved before the parent concludes.
For an SCFToken, the lifecycle is atomic:
\begin{equation}
	F:
	\Sigma_0
	\xrightarrow[\text{call}]{\text{SCFToken}}
	\Sigma_{\text{active}}
	\xrightarrow[\text{complete}]{\text{wait}\ F\ \text{done}}
	\Sigma_0
\end{equation}
A concrete tree-structured lifecycle example with nested XML elements is illustrated in Appendix~\ref{appendix:mapper} (Fig.~\ref{fig:xml-lifecycle}).
Detailed formal definitions are also provided therein.

\subsubsection{Scheduler}\label{sec:scheduler}

The scheduler receives mapped function calls and dispatches them to designated execution channels.
It adopts a multi-channel architecture consisting of one main channel and multiple sub-channels, where each sub-channel typically corresponds to a functional module (\eg, a specific limb, speech, or vision) that imposes intra-module mutual exclusion.
The scheduling obeys three rules:
\begin{itemize}
    \item \textbf{Intra-channel Synchronous Dispatch}: Functions on the same channel are dispatched synchronously, ensuring sequential execution within each component.
    \item \textbf{Inter-channel Asynchronous Dispatch}: Functions on different channels are dispatched asynchronously, enabling parallel execution across components (\eg, locomotion while speaking).
    \item \textbf{Global Blocking}: The main channel blocks all subsequent dispatches globally until its current function completes.
\end{itemize}
Formal definitions are provided in Appendix~\ref{appendix:scheduler}.

\subsubsection{Executor}\label{sec:executor}
The executor binds each scheduled function call to its concrete, platform-specific implementation at runtime.
On the hardware side, physical actuators such as servo motors are driven through low-level control processes (\eg, ROS~\cite{quigley2009ros} nodes or embedded firmware handlers).
On the software side, modules such as text-to-speech engines, audio players, and screen animation renderers are invoked as callable services.
While the underlying implementations vary across devices, operating systems, and programming languages, they all conform to the same abstract function interfaces defined in Section~\ref{sec:code-as-prompt-main}, enabling \algname to operate portably across heterogeneous platforms.

After each function completes, the executor captures results or exceptions and returns them as observations.
This supports ReAct-style reasoning loops~\cite{yao2023react}, where the LLM interleaves reasoning with actions and receives observations from the environment to inform subsequent steps.

\section{Experiment}\label{sec:experiment}
We evaluate \algname on our robot prototype \rbtname (Fig.~\ref{fig:robot}) through three experiment groups: structured behavioral tasks, long-horizon multimodal scenarios, and an ablation study isolating the contribution of the \textit{function token} design.

\subsection{Evaluation Metric}

We introduce the \textbf{Directed Structured Behavior Correctness (DSBC)} metric to evaluate the structural and functional fidelity of a generated XML element tree $T$ against a ground truth tree $G$.
DSBC performs a layer-wise comparison that enforces hierarchical alignment while permitting directed attribute flexibility.

\textbf{Preliminaries.}
Let $L_l^G=(e_{l,1}^G, \dots, e_{l,m_l}^G)$ and $L_l^T=(e_{l,1}^T, \dots, e_{l,n_l}^T)$ denote the ordered element sequences at depth $l$ in $G$ and $T$, with lengths $m_l$ and $n_l$, respectively.
For each element $e$, let $\tau(e)$ denote its tag name and $A(e)$ its set of attribute key-value pairs.
Let $D^{*} = \max(d_G, d_T)$ be the maximum depth across both trees, where $d_G$ and $d_T$ denote the depths of $G$ and $T$. Any layer absent in either tree is treated as an empty sequence.

\textbf{Tag Sequence Score.}
At each layer $l$, we assess both tag count accuracy and positional order.
Let $\mathcal{T}_l$ denote the set of distinct tags in $L_l^G \cup L_l^T$, and let $N_\tau^G, N_\tau^T$ be the occurrence counts of tag $\tau$ in $L_l^G$ and $L_l^T$.
The count accuracy is
\begin{equation}
	C_{\mathrm{cnt}}(l) = 1 - \frac{\sum_{\tau \in \mathcal{T}_l} |N_\tau^G - N_\tau^T|}{\sum_{\tau \in \mathcal{T}_l} \max(N_\tau^G,\, N_\tau^T)}\,,
\end{equation}
defined as $1$ when the denominator vanishes (\ie, both sequences are empty).
Let $M_l = \max(m_l, n_l)$ and $\delta_i = \mathbf{1}[\tau(e_{l,i}^G) = \tau(e_{l,i}^T)]$ be the element-wise tag match indicator. The positional order accuracy is
\begin{equation}
	C_{\mathrm{ord}}(l) = \frac{1}{M_l} \sum_{i=1}^{\min(m_l,\, n_l)} \delta_i\,,
\end{equation}
which jointly penalizes positional mismatches and length discrepancies.
The combined tag score at layer $l$ is $S_{\mathrm{tag}}(l) = (C_{\mathrm{cnt}}(l) + C_{\mathrm{ord}}(l))/2$.

\textbf{Attribute Score.}
For each ground truth element $e^G$ at layer $l$, we define the set of tag-compatible generated elements $\mathcal{E}(e^G) = \{e^T \in L_l^T \mid \tau(e^T) = \tau(e^G)\}$ and compute the best-match attribute overlap:
\begin{equation}
	\phi(e^G) =
	\begin{cases}
		1                                                                                  & \text{if } A(e^G) = \emptyset,           \\
		0                                                                                  & \text{if } \mathcal{E}(e^G) = \emptyset, \\
		\displaystyle\max_{e^T \in \mathcal{E}(e^G)} \frac{|A(e^G) \cap A(e^T)|}{|A(e^G)|} & \text{otherwise.}
	\end{cases}
\end{equation}
The layer attribute score averages over all ground truth elements:
\begin{equation}
	S_{\mathrm{attr}}(l) =
	\begin{cases}
		1                                                           & \text{if } m_l = 0, \\
		\displaystyle\frac{1}{m_l} \sum_{i=1}^{m_l} \phi(e_{l,i}^G) & \text{otherwise.}
	\end{cases}
\end{equation}

\textbf{Final Score.}
The per-layer score $S_l = (S_{\mathrm{tag}}(l) + S_{\mathrm{attr}}(l))/2$ is averaged across all $D^{*}$ layers:
\begin{equation}
	\mathrm{DSBC}(G, T) = \frac{1}{D^{*}} \sum_{l=0}^{D^{*}-1} S_l \;\in [0,\,1]\,,
\end{equation}
where $1$ indicates perfect structural and attributive correspondence.
$\mathrm{DSBC}$ is defined as $1$ when $D^{*}=0$.

\subsection{Experimental Setup}
We design three experiment groups to evaluate single-turn behavioral generation, multi-turn multimodal reasoning, and the contribution of \textit{function tokens} over native function calls, respectively.

\textbf{Grounded HRI Tasks.}
We construct 30 grounded HRI tasks spanning five behavioral modes of increasing complexity: atomic (3), sequential (10), parallel (4), sequential-parallel composite (5), and lifecycle-scoped termination (8), with mode definitions in Appendix~\ref{appendix:behavioral-patterns} and task descriptions in Appendix~\ref{appendix:task-descriptions}.
Each task is evaluated using the DSBC metric with five independent trials per model-task combination to account for generation nondeterminism.
All models are queried through their official APIs with default decoding parameters (\ie, no custom \texttt{temperature}, \texttt{top\_p}, or other sampling overrides).
We evaluate 21 LLMs from nine providers (OpenAI, Anthropic, Google, \etc) and select the best-performing model from each provider for comparison in Table~\ref{tab:comprehensive_results} (full results in Appendix~\ref{appendix:detailed-scores}).
Reasoning models are excluded, as their extended chain-of-thought latency is incompatible with the real-time responsiveness required for embodied interaction.

\begin{figure}[ht]
	\centering%
	\subfigure[Bottle Collision]{%
		\includegraphics[width=0.32\columnwidth]{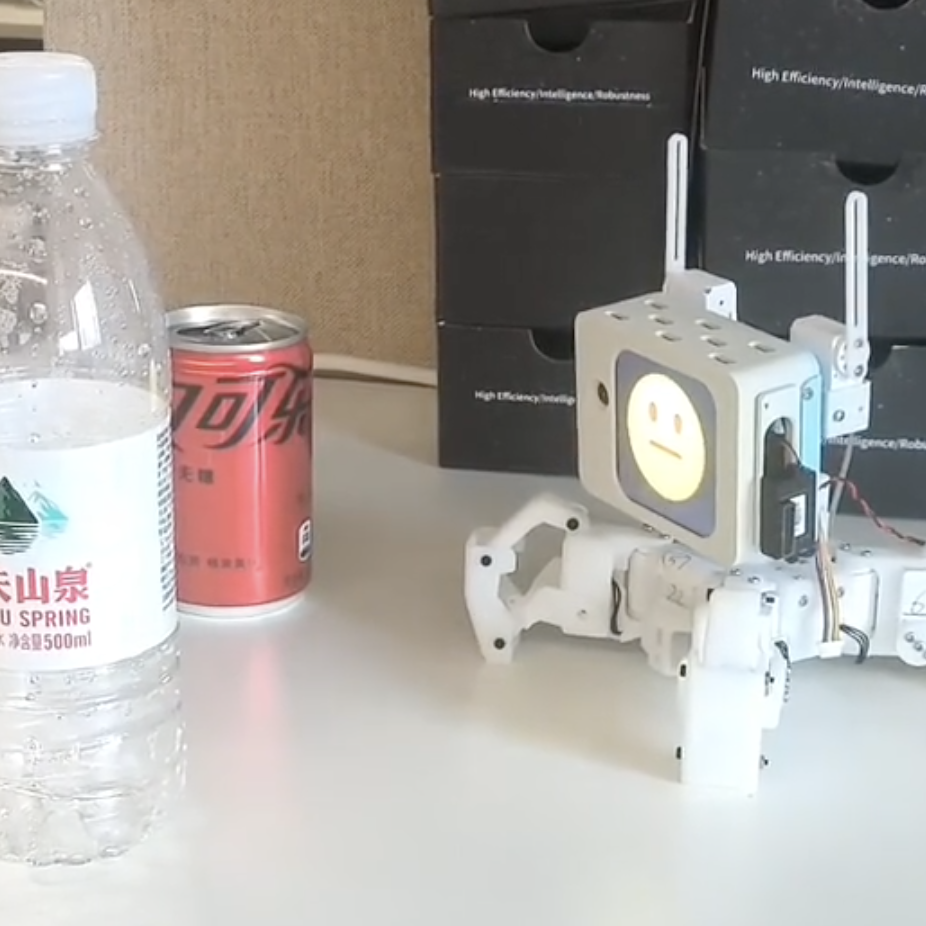}%
		\label{fig:long-bottle}%
	}\hfill%
	\subfigure[Mirror Test]{%
		\includegraphics[width=0.32\columnwidth]{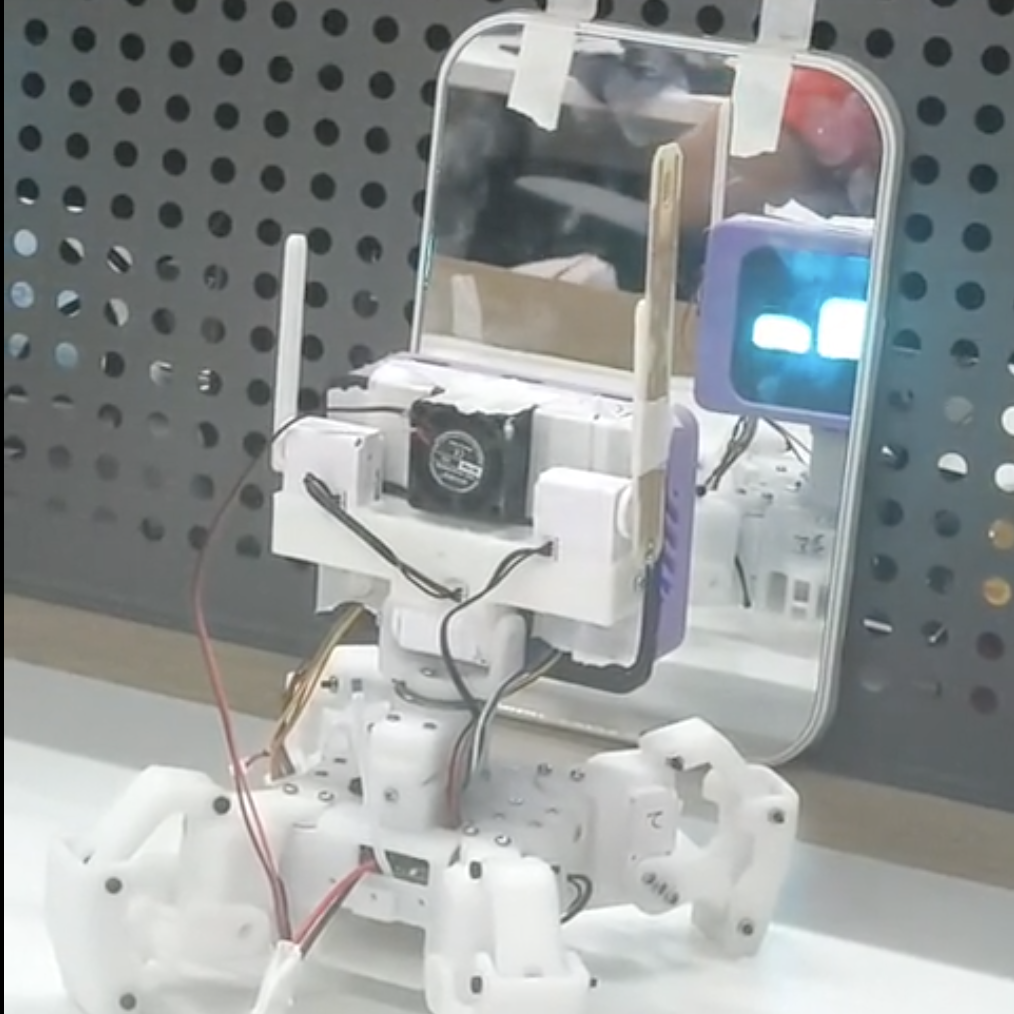}%
		\label{fig:long-mirror}%
	}\hfill%
	\subfigure[RPS Game]{%
		\includegraphics[width=0.32\columnwidth]{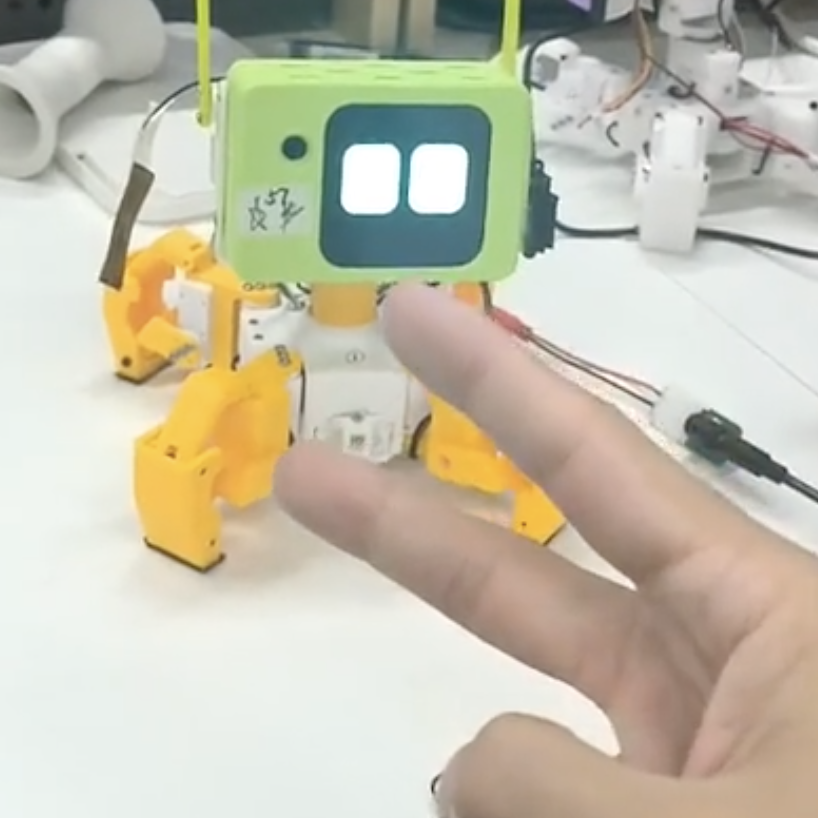}%
		\label{fig:long-rps}%
	}
	\caption{Three long-horizon multimodal task experiments.}
	\label{fig:long-horizon-tasks}
\end{figure}

\textbf{Long-Horizon Multimodal Tasks.}
We further evaluate \algname on three long-horizon multimodal tasks (Fig.~\ref{fig:long-horizon-tasks}) that leverage the ReAct-based reasoning loop (Section~\ref{sec:executor}):
(1) \textit{bottle collision}, where the robot must orient itself toward a bottle, walk forward to collide with it, and visually verify whether the bottle has been toppled,
(2) \textit{mirror self-recognition}, where two robots are placed before a mirror and the tested robot must identify which reflection is its own by performing deliberate actions and observing corresponding motion, and
(3) \textit{rock-paper-scissors}, where the robot plays against a human opponent by verbally declaring its choice, visually recognizing the human's hand gesture, and determining the winner of each round.
Each task involves open-ended multi-turn interactions where the behavioral sequences cannot be predetermined, making ground truth XML trees unavailable for DSBC evaluation.
We therefore adopt a human evaluation protocol with task-specific rubrics on a 10-point scale (Appendix~\ref{appendix:long-horizon-scoring}).
Each trial is independently scored by three evaluators and averaged, with three trials conducted per model-task combination.
Three vision-capable LMMs that support the visual reasoning loop are evaluated: GPT-4.1, Claude-Sonnet-4, and Doubao-Seed-1.6.
Full task instructions are provided in Appendix~\ref{appendix:task-descriptions} (Tasks 31--33).

\textbf{Ablation: Function Tokens \vs Native Function Calls.}
To isolate the contribution of our \textit{function token} design, we conduct an ablation replacing it with the native function call protocol (\eg, OpenAI tool calls) while keeping all other system components identical.
This comparison is motivated by a fundamental architectural difference: native function calls separate textual content and tool invocations into disjoint API response fields and represent calls as a flat JSON array without hierarchical nesting.
We independently design 15 tasks with three per behavioral mode (atomic, sequential, parallel, sequential-parallel composite, and lifecycle-scoped termination), separate from the 30 grounded HRI tasks, to systematically assess whether native function calls can express the structural patterns required by each mode.
Each task is evaluated with a binary pass/fail criterion using GPT-4.1.

\subsection{Results}

\subsubsection{Grounded HRI Tasks}

\begin{table*}[t]
	\centering
	\small
	\caption{DSBC scores across five behavioral modes for the top 9 of 21 evaluated LLMs (5 trials per model-task pair).}
	\begin{tabular*}{\textwidth}{@{\extracolsep{\fill}}lccccccc}
		\toprule
		\textbf{Model} & Atomic & Sequential & Parallel & Seq.-Par. & Lifecycle-Scoped & \textbf{Micro} & \textbf{Macro} \\
		& & & & Composite & Termination & \textbf{($\pm$ Std)} & \textbf{($\pm$ Std)} \\
		\midrule
		Claude-Sonnet-4 & \textbf{1.00} & \textbf{0.81} & 0.80 & \textbf{0.79} & \textbf{0.83} & \textbf{0.83} ($\pm$ 0.02) & \textbf{0.84} ($\pm$ 0.01) \\
		GPT-4.1 & \textbf{1.00} & 0.68 & 0.83 & 0.64 & 0.73 & 0.74 ($\pm$ 0.01) & 0.78 ($\pm$ 0.02) \\
		Doubao-Seed-1.6 & \textbf{1.00} & 0.66 & 0.88 & 0.77 & 0.56 & 0.71 ($\pm$ 0.01) & 0.77 ($\pm$ 0.01) \\
		Kimi-K2 & \textbf{1.00} & 0.68 & 0.84 & 0.72 & 0.55 & 0.70 ($\pm$ 0.02) & 0.76 ($\pm$ 0.03) \\
		Grok-3 & \textbf{1.00} & 0.70 & \textbf{0.89} & 0.78 & 0.81 & 0.80 ($\pm$ 0.03) & \textbf{0.84} ($\pm$ 0.03) \\
		Gemini-2.5-Flash & \textbf{1.00} & 0.73 & 0.73 & 0.72 & 0.80 & 0.78 ($\pm$ 0.01) & 0.80 ($\pm$ 0.01) \\
		Qwen3-Max & \textbf{1.00} & 0.64 & 0.68 & 0.65 & 0.79 & 0.72 ($\pm$ 0.01) & 0.75 ($\pm$ 0.01) \\
		Llama-3.2-90B & \textbf{1.00} & 0.67 & 0.77 & 0.60 & 0.60 & 0.69 ($\pm$ 0.04) & 0.73 ($\pm$ 0.04) \\
		DeepSeek-V3.1 & 0.90 & 0.65 & 0.68 & 0.68 & 0.60 & 0.67 ($\pm$ 0.03) & 0.70 ($\pm$ 0.03) \\
		\bottomrule
	\end{tabular*}
	\label{tab:comprehensive_results}
\end{table*}

Table~\ref{tab:comprehensive_results} reports DSBC scores for the top nine of 21 evaluated models (full per-task breakdown in Appendix~\ref{appendix:detailed-scores}).
Claude-Sonnet-4 achieves the highest micro-average (task-weighted, \textbf{0.83}$\pm$0.02), followed by Grok-3 (0.80), Gemini-2.5-Flash (0.78), and GPT-4.1 (0.74).
Claude-Sonnet-4 and Grok-3 share the top macro-average (equal-weight across modes) of \textbf{0.84}, and all models except DeepSeek-V3.1 score perfectly on atomic tasks.

Mode-level analysis reveals distinct capability profiles.
On sequential tasks, Claude-Sonnet-4 leads (0.81) by a wide margin over Gemini-2.5-Flash (0.73).
For parallel execution, Grok-3 (0.89) and Doubao-Seed-1.6 (0.88) rank highest, while Qwen3-Max and DeepSeek-V3.1 trail at 0.68.
Lifecycle-scoped termination exhibits the highest cross-model variance: Claude-Sonnet-4 (0.83), Grok-3 (0.81), and Gemini-2.5-Flash (0.80) remain strong, yet Kimi-K2 (0.55) and Doubao-Seed-1.6 (0.56) drop sharply despite strong parallel results (0.84 and 0.88), suggesting that flat concurrency and nested lifecycle management demand distinct capabilities.
Across all nine models, atomic tasks achieve a near-perfect model-averaged score of 0.99, while all other modes range between 0.69 (sequential) and 0.79 (parallel), and no single model dominates all modes, confirming that structured behavioral generation remains a significant open challenge.

\subsubsection{Long-Horizon Multimodal Tasks}

\begin{table}
	\centering
	\caption{Human evaluation scores for long-horizon multimodal tasks (10-point scale, averaged over 3 trials).}
	\begin{tabular}{lccc}
		\toprule
		\textbf{Task}    & \textbf{GPT-4.1} & \textbf{Claude-Sonnet-4} & \textbf{Doubao-Seed-1.6} \\
		\midrule
		Bottle Collision & 8.7              & 8.0                      & 5.3                      \\
		Mirror Test      & 3.3              & 3.3                      & 1.7                      \\
		RPS Game         & 9.0              & 9.3                      & 9.3                      \\
		\midrule
		\textbf{Avg.}    & \textbf{7.0}     & 6.9                      & 5.4                      \\
		\bottomrule
	\end{tabular}
	\label{tab:long_horizon}
\end{table}

Table~\ref{tab:long_horizon} reports human evaluation scores averaged over three trials (per-trial breakdown in Appendix~\ref{appendix:long-horizon-scores}).
All models score highly on rock-paper-scissors ($\geq$9.0) and bottle collision (GPT-4.1: 8.7, Claude-Sonnet-4: 8.0), confirming reliable visual recognition and rule-based reasoning.
Mirror self-recognition yields universally low scores (GPT-4.1 and Claude-Sonnet-4: 3.3, Doubao-Seed-1.6: 1.7), as models tend to draw conclusions without performing active motion-based verification (Appendix~\ref{appendix:error-analysis}).
GPT-4.1 achieves the highest average of \textbf{7.0}, followed by Claude-Sonnet-4 (6.9), while Doubao-Seed-1.6 trails at 5.4 with notably lower scores in both bottle collision (5.3 \vs 8.0--8.7) and mirror self-recognition (1.7 \vs 3.3).

\subsubsection{Function Tokens \vs Native Function Calls}

\begin{table*}[t]
	\centering
	\caption{Function tokens vs.\ native function calls on behavioral mode completion (GPT-4.1).}
	\begin{tabular*}{\textwidth}{@{\extracolsep{\fill}}l c l cc@{}}
		\toprule
		\textbf{Behavioral Mode}                                  & \textbf{\#} & \textbf{Tasks}                                                                       & \multicolumn{2}{c}{\textbf{Task Completion \scriptsize{(GPT-4.1)}}}                         \\
		\cmidrule(lr){4-5}
		& &                                                                                      & \textbf{Func.\ Token}                                                          & \textbf{Native Call} \\
		\midrule
		\multirow{3}{*}{\textbf{Atomic}}                         & 1 & \scriptsize\textit{Say hello}                                                        & \cmark                                                             & \cmark       \\
		& 2 & \scriptsize\textit{Nod your head for 3 seconds}                                      & \cmark                                                             & \cmark       \\
		& 3 & \scriptsize\textit{Do 3 push-ups}                                                    & \cmark                                                             & \cmark       \\
		\addlinespace
		\multirow{3}{*}{\textbf{Sequential}}                     & 4 & \scriptsize\textit{Take a step forward, then shake your head}                            & \cmark                                                             & \cmark       \\
		& 5 & \scriptsize\textit{Take a step forward, then rotate 90 degrees, then take another step forward} & \cmark                                                             & \xmark       \\
		& 6 & \scriptsize\textit{Say hello, then take a step forward, then say goodbye}             & \cmark                                                             & \xmark       \\
		\addlinespace
		\multirow{3}{*}{\textbf{Parallel}}                       & 7 & \scriptsize\textit{Take a step forward while shaking your head}                                        & \cmark                                                             & \cmark       \\
		& 8 & \scriptsize\textit{Take a step forward while shaking your head and moving your ears}                         & \cmark                                                             & \cmark       \\
		& 9 & \scriptsize\textit{Take a step forward while saying hello}                                             & \cmark                                                             & \xmark       \\
		\addlinespace
		\multirow{3}{*}{\textbf{Seq.-Par. Composite}}           & 10 & \scriptsize\textit{Take a step forward, then rotate while shaking your head}          & \cmark                                                             & \xmark       \\
		& 11 & \scriptsize\textit{Smile while wiggling your ears, then step backward while saying ``stepping back''} & \cmark                                                             & \xmark       \\
		& 12 & \scriptsize\textit{Say hello, then rotate while shaking your head, then say goodbye}  & \cmark                                                             & \xmark       \\
		\addlinespace
		\multirow{3}{*}{\textbf{Lifecycle-Scoped Term.}}   & 13 & \scriptsize\textit{Keep rotating while shaking your ears 3 times, stop rotating after the ear shaking is done}                & \cmark                                                             & \xmark       \\
		& 14 & \scriptsize\textit{Keep rotating while saying hello 3 times, stop rotating after greeting is done}                    & \cmark                                                             & \xmark       \\
		& 15 & \scriptsize\textit{Keep rotating while shaking your head and ears, stop rotating after both are done}      & \cmark                                                             & \xmark       \\
		\bottomrule
	\end{tabular*}
	\label{tab:function_call}
\end{table*}

Table~\ref{tab:function_call} reports task completion across all five behavioral modes.
\textit{Function tokens} achieve perfect completion (15/15), whereas native function calls succeed on only 6 of 15 tasks.
The six successful cases (Tasks~1--4, 7, 8) involve at most two-step sequences or flat unimodal parallelism.
All nine failures stem from three structural limitations of the native function call interface, with individual tasks potentially exhibiting multiple limitations.

\textbf{Compositional action loss.}
As task structure exceeds two-step sequences or flat parallelism, native calls progressively drop constituent actions.
This affects Task~5 (three-step sequential) and all sequential-parallel composite tasks (Tasks~10--12), with Tasks~11 and~12 further compounded by linguistic-behavioral decoupling.

\textbf{Linguistic-behavioral output decoupling.}
The native API routes text and tool invocations through disjoint response fields, preventing speech and motor actions from co-occurring.
All tasks requiring speech-motor coordination (Tasks~6, 9, 11, 12, and~14) lose their speech components regardless of task complexity.

\textbf{Topological inexpressibility.}
Native calls encode invocations as a flat JSON array without mechanisms for nesting or lifecycle scoping.
Lifecycle-scoped termination requires a parent action to persist until child completion, a dependency unrepresentable in a flat call list.
All three lifecycle-scoped tasks (Tasks~13--15) fail entirely, with Task~14 additionally exhibiting linguistic-behavioral decoupling.

Our XML-based \textit{function tokens} resolve these limitations through streaming action programming, where textual and functional tokens are interleaved within a single autoregressive stream, enabling the model to emit speech and structured actions as a unified, incrementally executable program with hierarchically nested topology.
Detailed per-task output analysis is provided in Appendix~\ref{appendix:funcall-analysis}.

\section{Conclusion}\label{sec:conclusion}
In this work, we introduce \algname, featuring a \textit{function token} schema that enables streaming function calls from LLMs and a multi-channel scheduling algorithm for concurrent behavioral programming in multi-component embodied systems.
Our key insight lies in enabling human-like embodied interaction through on-the-fly execution and coordinated multi-component actions.
Experimental evaluation on robot \rbtname across 30 grounded HRI tasks shows that Claude-Sonnet-4 achieves the highest DSBC of \textbf{0.83}, while long-horizon multimodal tasks reach a best human evaluation score of \textbf{7.0}/10 with GPT-4.1.
Crucially, our \textit{function token} schema achieves complete task execution across all five behavioral modes, whereas native function calls fail on tasks requiring linguistic-behavioral coordination or hierarchical nesting, particularly lifecycle-scoped termination.
Beyond physical robotics, our approach generalizes to embodied conversational agents, mixed-reality interaction systems, and LLM-based autonomous agents.
Limitations and future work are discussed in Appendix~\ref{appendix:limitations}.

\ifblindreview\else
  \section*{Acknowledgment}
  The name of our approach pays tribute to the science fiction work \textit{Ghost in the Shell} by Masamune Shirow.
  We appreciate all those who strive to turn science fiction into reality.
\fi

\bibliographystyle{IEEEtran}
\bibliography{ref}

\ifblindreview\else
  \clearpage

  \appendices

  \section{\algname Implementation Details}\label{appendix:implementation}

In \algname, the LLM performs declarative embodied programming.
Its autoregressive generation produces structured XML-encoded function calls that the shell runtime incrementally parses, maps to concrete interfaces, schedules across execution channels, and executes on the target platform.
The XML function token schema (Section~\ref{appendix:xml-schema}) specifies the encoding syntax governing how function calls are expressed, abstract function interfaces (Section~\ref{sec:code-as-prompt}) define the set of callable operations available to the LLM, and the shell pipeline (Sections~\ref{appendix:parser}--\ref{appendix:mapper}) transforms the encoded output into executable function invocations through streaming parsing, mapping, and scheduling.

\subsection{XML Function Token Schema Specification}\label{appendix:xml-schema}

\textbf{Activation Function Token (AFToken):} Corresponds to the \textit{start-tag} in XML.
An AFToken transitions a function from its initial state to an active execution state and persists until explicitly reset.
Syntactically, it begins with \texttt{<}, followed by a function name, optional whitespace and attributes (parameters), and ends with \texttt{>}.
Formally, $\text{AFToken} ::= \texttt{<}\ \text{FName}\ (\text{S}\ \text{FParam})^*\ \text{S}?\ \texttt{>}$, where $\text{FName}$ is the function name and $\text{FParam}$ denotes a function parameter.

\textbf{Reset Function Token (RFToken):} Corresponds to the \textit{end-tag} in XML.
An RFToken resets the function from its active state back to the initial state, thereby completing the function lifecycle.
It consists of \texttt{</}, the function name, optional whitespace, and \texttt{>}.
Formally, $\text{RFToken} ::= \texttt{</}\ \text{FName}\ \text{S}?\ \texttt{>}$.

\textbf{Self-Contained Function Token (SCFToken):} Corresponds to the \textit{empty-element tag} in XML.
An SCFToken encodes an atomic, self-contained function invocation with no nested tokens.
Its lifecycle terminates immediately upon completion.
Syntactically, it consists of \texttt{<}, a function name, optional parameters, and \texttt{/>}.
Formally, $\text{SCFToken} ::= \texttt{<}\ \text{FName}\ (\text{S}\ \text{FParam})^*\ \text{S}?\ \texttt{/>}$.

\textbf{Function Character Token (FChar):} Corresponds to the \textit{character data} in XML.
FChar represents literal character data within the encoding and can serve as either direct output or long-form function input.
FChar must not contain the characters \texttt{<}, \texttt{\&}, or the sequence \texttt{]]>}.
Formally, $\text{FChar} ::= \left[\ \hat{}\ \texttt{<\&}\ \right]^* - \left(\left[\ \hat{}\ \texttt{<\&}\ \right]^*\ \texttt{]]>}\ \left[\ \hat{}\ \texttt{<\&}\ \right]^*\right)$.
When the reserved characters (\eg, \texttt{<}, \texttt{\&}) are required, they must each be encoded as a \textbf{Function Reference Token (FRef)}, which corresponds to a \textit{reference} in XML.
An FRef begins with \texttt{\&} and ends with \texttt{;}, enabling the representation of predefined entities or character references within parameter values or textual content.

\textbf{Parameter Name (FParamName):} Corresponds to the XML \textit{attribute name}.
It denotes the identifier of a parameter within an AFToken or SCFToken and must satisfy the naming rules specified in the XML standard.

\textbf{Parameter Value (FParamValue):} Corresponds to the XML \textit{attribute value}.
It denotes the value assigned to a parameter name, enclosed in single or double quotes, following the constraints on \textit{attribute values} in XML.

\textbf{Function Parameter (FParam):} Corresponds to the XML \textit{attribute}.
It is composed of a parameter name and its assigned value.
Formally, $\text{FParam} ::= \text{FParamName}\ \text{Eq}\ \text{FParamValue}$.

In the above definitions, $\text{S}$ denotes one or more whitespace characters as defined in XML, and FName follows the XML \textit{Name} production rule.
Terms not explicitly defined here follow the XML 1.0 specification~\cite{bray2000extensible}.
Both the LLM system prompt and the shell runtime must strictly conform to this encoding specification.

\subsection{Code as Prompt}\label{sec:code-as-prompt}

This subsection elaborates on the code abstraction design introduced in Section~\ref{sec:code-as-prompt-main}.
In \algname, the LLM's system prompt comprises two components: (1)~the XML function token specification (Section~\ref{appendix:xml-schema}), which defines the encoding syntax for generating function calls, and (2)~a set of abstract function interfaces exported by the target device, which enumerate the currently available operations.
Together, they enable the LLM to perceive the system's capabilities and generate well-formed control encodings according to the specification, without any knowledge of the underlying implementation.

\subsubsection{Interface Representation}

Each abstract function interface declares
(1)~a function name identifying a distinct capability,
(2)~typed parameters with annotations (\eg, \texttt{Literal}, \texttt{float}, \texttt{bool}), default values, and value range constraints, and
(3)~a documentation comment describing the function semantics and parameter usage.
All implementation bodies are omitted, and only the declarative signature is retained.

Two equivalent representation formats are supported.
In the \textit{code-based format}, each capability is expressed as a function stub in the host language (\eg, Python \textit{def} with type hints and docstring, or JavaScript with \textit{JSDoc} annotations).
For example, a head action interface declares the action type as a \texttt{Literal} parameter constrained to specific values, alongside a \texttt{float} duration parameter with a defined valid range.
In the \textit{XML-based format}, the same capability is encoded as a \textit{function token} (Section~\ref{sec:xml-schema}), with parameter specifications embedded as attributes and semantic descriptions provided in XML comments.
Both formats convey identical declarative information, and the choice is transparent to the shell pipeline.

\subsubsection{Unified Abstraction of Hardware and Software}

The abstraction uniformly covers both hardware and software capabilities.
Locomotion primitives such as \texttt{stand}, \texttt{move}, and \texttt{rotate} encapsulate low-level servo control driven by ROS nodes or embedded firmware handlers, while \texttt{speak} and \texttt{play\_bgm} wrap software services such as text-to-speech engines and audio players.
Both types are exposed to the LLM as declarative function interfaces of the same form, and no distinction between hardware and software is visible at the prompt level.

\subsubsection{Dynamic Module Composition}

At runtime, each abstract interface is bound to its concrete implementation via dependency injection (DI).
Functional modules, including physical devices (\eg, motor controllers, cameras) and software components (\eg, audio players, vision models), can be dynamically mounted and unmounted.
Their corresponding function interfaces are automatically reflected in or removed from the system prompt.
Consequently, the LLM adapts to a varying set of available capabilities across multi-turn interactions without retraining or manual prompt modification.

\subsubsection{Device-Agnostic Generality}

Because \algname requires only the encoding specification and a set of abstract function interfaces, any device or software system that exports conforming interfaces can be driven by an LLM through the shell pipeline.
The approach is therefore independent of any specific programming language, deployment scenario, or model architecture.
Downstream implementations can integrate heterogeneous execution backends such as reinforcement learning (RL), imitation learning (IL), vision-language-action (VLA) models, and classical motion planners, as long as they conform to the declared function signatures.
This generality extends beyond robotics to domains such as digital human animation, game engine control, and operating system automation.

\subsection{Shell Parser}\label{appendix:parser}

Formally, let $O = [o_1, o_2, \ldots, o_n]$ denote the output token sequence generated by the LLM.
The parser processes $O$ in generation order and incrementally transforms it into a structured sequence $S$ of \textit{function tokens}:
\begin{multline}
	\mathcal{P}(O) \rightarrow S = [s_1, s_2, \ldots, s_l],\\
	s_i \in \{\text{AFToken},\ \text{RFToken},\ \text{SCFToken},\ \text{FChar},\ \text{FRef}\}
\end{multline}
where $\mathcal{P}(\cdot)$ is the parsing function, and each $s_i$ denotes a parsed \textit{function token}.

During the parsing process, when specific token patterns are identified, they form complete XML elements.
An element $E$ is constructed when:
\begin{multline}
	E ::= \text{FChar} \mid \text{FRef} \mid \text{SCFToken}\\
	\mid \left(\text{AFToken} \cdot \text{Elements} \cdot \text{RFToken}\right)
\end{multline}
where $\text{Elements} ::= [E_1, E_2, \ldots, E_k]$ denotes a (possibly empty) sequence of nested elements, with $k \geq 0$.

For non-empty elements of the form $\text{AFToken} \cdot \text{Elements} \cdot \text{RFToken}$, a strict parsing constraint is imposed: AFToken and RFToken must have identical FName.
If this condition is violated, a parsing exception is raised and processing is immediately terminated.

\subsection{Shell Mapper}\label{appendix:mapper}

\textbf{Systematic Mapping Overview:}
For each parsed XML element $E$, the mapper defines a deterministic mapping to the tuple $(F, \vec{\theta})$, where $F$ denotes the function interface resolved from $E$'s FName, and $\vec{\theta}$ is the vector of typed parameters obtained through attribute-to-parameter conversion:
\begin{equation}
	\mathcal{M}(E)=(F,\ \vec{\theta})
\end{equation}
The execution of $F(\vec{\theta})$ then strictly follows the lifecycle prescribed by $\mathcal{L}(E)$.
We detail the two mappings below.

\textbf{Attribute-to-Parameter Mapping:}
When an XML element $E$ contains a list of FParams $(a_k, v_k)$, each parameter must be aligned with the function interface's type signature.
We introduce a type conversion operator $T(\cdot)$ to ensure each FParamValue is converted to the corresponding parameter type of $F$:
\begin{multline}
	E(a_1=v_1,\ \ldots,\ a_n=v_n) \implies \\
	F\left(\theta_1=T(v_1),\ \ldots,\ \theta_n=T(v_n)\right)
\end{multline}
where $a_k$ is an FParamName, $v_k$ is its FParamValue, and $T(v_k)$ converts the raw data into the type required by $\theta_k$ in the target function.

As shown in Fig.~\ref{fig:xml-map-func}, we illustrate the mapping with \texttt{move} and \texttt{smile} as examples.
The FParams \texttt{speed}, \texttt{size}, and \texttt{time} are mapped to function arguments in different programming languages.
During mapping, string-typed FParamValues are cast to the target types required by the host language, \eg, \texttt{speed} is converted from a string to a float.

\textbf{Element-to-Function Lifecycle Mapping:}
We formalize the execution lifecycle of an XML element as a state transition process, denoted by $\mathcal{L}$.
For a non-empty element $E$ with mapped function $F$, the lifecycle is:
\begin{multline}
	\mathcal{L}(E) \to F:\\
	\Sigma_0
	\xrightarrow[\text{activate}]{\text{AFToken}}
	\Sigma_{\text{active}}
	\xrightarrow[\text{execute}]{\{\mathcal{L}(E_i)\}_{i=1}^k}
	\Sigma_{\text{active}}
	\xrightarrow[\text{reset}]{\text{RFToken}}
	\Sigma_0
	\label{eq:element-lifecycle}
\end{multline}
where $\Sigma_0$ is the initial state, $\Sigma_{\text{active}}$ is the active execution state, and $\{E_i\}_{i=1}^k$ ($k \ge 0$) is the (possibly empty) sequence of child elements nested within $E$, each recursively mapped by $\mathcal{L}(E_i)$ to its corresponding function call.
The term $\big\{\mathcal{L}(E_i)\big\}_{i=1}^k$ explicitly represents tree-structured recursive execution.

For an SCFToken element, the lifecycle is a single-step atomic invocation:
\begin{multline}
	\mathcal{L}(E) \to F:\\
	\Sigma_0
	\xrightarrow[\text{call}]{\text{SCFToken}}
	\Sigma_{\text{active}}
	\xrightarrow[\text{complete}]{\text{wait}\ F\ \text{done}}
	\Sigma_0
\end{multline}

Fig.~\ref{fig:xml-lifecycle} illustrates a tree-structured example of nested XML elements and their corresponding function executions.
In this diagram, AF denotes AFToken, RF denotes RFToken, and the dashed arrows indicate the order of state transitions during execution.
The green-shaded area highlights the execution lifecycle of $F_0$, while the yellow-shaded region encapsulates the lifecycle of $F_{11}$.
Both $F_{21}$ and $F_{22}$ are SCFToken elements, hence their lifecycles terminate atomically upon completion.

\begin{figure}[ht]
	\centering
	\includegraphics[width=\columnwidth]{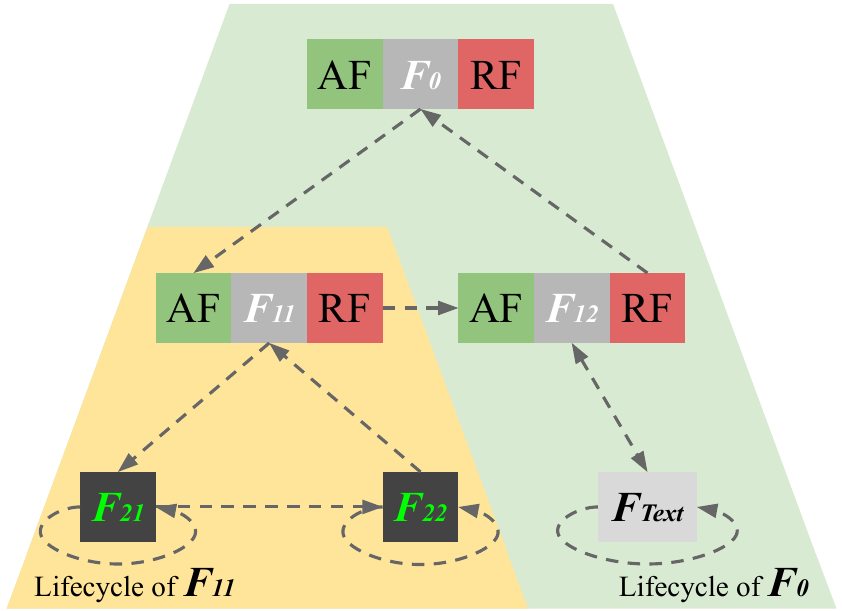}
	\caption{A tree-structured example of nested XML elements and their corresponding function lifecycles. The green-shaded area shows the lifecycle of $F_0$, and the yellow-shaded region shows $F_{11}$.}
	\label{fig:xml-lifecycle}
\end{figure}

Whether each function's state transitions are executed synchronously or asynchronously is determined by its assigned scheduling channel, as discussed in Section~\ref{sec:scheduler}.

\subsection{Shell Scheduler}\label{appendix:scheduler}

Let the function call queue output from the mapper be denoted as an ordered sequence $\mathcal{F} = [F_1, F_2, \dots, F_n]$, where each $F_i$ represents a function call.
Let $\mathcal{C} = \{ C_0, C_1, \dots, C_m \}$ be the set of execution channels, where $C_0$ denotes the main channel and $C_k$ ($k \geq 1$) are sub-channels.

The channel assignment is defined by the mapping function:
\begin{equation}
	\alpha: \mathcal{F} \rightarrow \mathcal{C},\;
	\alpha(F_i) = C_k,\;
	F_i \in \mathcal{F},\;
	C_k \in \mathcal{C}
\end{equation}

This guarantees each function $F_i$ is assigned to exactly one execution channel.

To formalize the scheduling semantics within and across channels, we define two scheduling relations:
\begin{itemize}
	\item $\Gamma(F_i, F_j)$: $F_i$ and $F_j$ execute synchronously, \ie, $F_j$ waits for $F_i$ to complete before starting.
	\item $\Lambda(F_i, F_j)$: $F_i$ and $F_j$ execute asynchronously, \ie, both $F_i$ and $F_j$ may proceed in parallel.
\end{itemize}

For all $F_i, F_j \in \mathcal{F}$, the scheduling rules governing function execution within the channel architecture are as follows:

Functions assigned to the \textbf{same channel} are scheduled synchronously and executed sequentially:
\begin{equation}
	\alpha(F_i) = \alpha(F_j) \wedge (i < j)
	\;\Rightarrow\;
	\Gamma(F_i, F_j)
\end{equation}

Functions assigned to \textbf{different channels} are scheduled asynchronously and may execute in parallel:
\begin{equation}
	\alpha(F_i) \neq \alpha(F_j)
	\;\Rightarrow\;
	\Lambda(F_i, F_j)
\end{equation}

If a function $F_i$ is assigned to the main channel $C_0$, all subsequent function calls $F_j$ with $j > i$ are globally blocked until $F_i$ has completed.
\begin{equation}
	\alpha(F_i) = C_0 \wedge (j > i)
	\;\Rightarrow\;
	\Gamma(F_i, F_j)
	\label{eq:main_channel_blocking}
\end{equation}

Based on the above foundations, the scheduler supports two complementary scheduling modes that differ in their default channel assignment policy.
We detail each mode below.

\subsubsection{Mode I: Async-First (Parallel-Priority)}\label{appendix:async-first}

In Async-First mode, each function $F_i$ is by default assigned to the sub-channel $C_k$ corresponding to its associated functional module (\eg, a specific limb, speaker, or camera).
Formally, let $\mu: \mathcal{F} \to \{C_1, \dots, C_m\}$ denote the module-to-channel mapping. The default assignment is:
\begin{equation}
	\alpha_{\text{async}}(F_i) = \mu(F_i), \quad \forall F_i \in \mathcal{F}
\end{equation}
Under this policy, functions on different modules execute concurrently by default, and synchronous ordering must be explicitly enforced by reassigning selected functions to the main channel $C_0$.

\textbf{This is the scheduling mode adopted throughout our paper.}
All 30 grounded HRI evaluation tasks and their ground truth \textit{function token} sequences in Table~\ref{tab:ground-truth} are designed under this mode.

\textbf{Synchronous Override via \texttt{<wait>}:}
To enforce sequential ordering in Async-First mode, we implement a dedicated function $F_{\text{wait}}$ corresponding to the \texttt{<wait>\ldots</wait>} \textit{function tokens}.
The XML element $E_{\text{wait}}$ conforms to the following structure:
\begin{equation}
	E_{\text{wait}} ::= \text{AFToken}_{\text{wait}} \cdot \text{Elements}_{\text{wait}}
	\cdot \text{RFToken}_{\text{wait}}
\end{equation}
where $\text{Elements}_{\text{wait}}$ denotes all sub-elements nested within $E_{\text{wait}}$.
Crucially, the sub-elements within $E_{\text{wait}}$ are still scheduled under the Async-First policy, \ie, they are scheduled to their respective module sub-channels and may execute concurrently with each other.
The synchronization boundary is imposed only at the level of $E_{\text{wait}}$ itself.
According to Eq.~\ref{eq:element-lifecycle}, the lifecycle $\mathcal{L}(E_{\text{wait}}) \to F_{\text{wait}}$ blocks until all $\text{Elements}_{\text{wait}}$ have completed execution.
By setting $\alpha(F_{\text{wait}}) = C_0$, $F_{\text{wait}}$ is assigned to the main channel, which prevents any subsequent function $F_j$ ($j > i_{\text{wait}}$) from entering execution until $E_{\text{wait}}$ terminates:
\begin{equation}
	\alpha(F_{\text{wait}}) = C_0 \;\Rightarrow\; \forall F_j,\ j > i_{\text{wait}}: \Gamma(F_{\text{wait}}, F_j)
\end{equation}

For example, consider Task~4 (Sequential Command) and Task~14 (Parallel Command) in Table~\ref{tab:ground-truth}.
In Task~4, the \texttt{<wait>} element wraps the speech output and assigns it to the main channel $C_0$, globally blocking all subsequent functions until the speech completes.
Only after the blocking is released does the smile expression execute on its sub-channel.
In contrast, Task~14 omits \texttt{<wait>}, so both the speech and the smile are scheduled to their respective sub-channels and execute concurrently by default.

\subsubsection{Mode II: Sync-First (Sequential-Priority)}\label{appendix:sync-first}

In Sync-First mode, all functions are by default assigned to the main channel $C_0$, resulting in strictly sequential execution:
\begin{equation}
	\alpha_{\text{sync}}(F_i) = C_0, \quad \forall F_i \in \mathcal{F}
\end{equation}
Under this policy, all functions execute sequentially by default, and concurrent execution must be explicitly requested.

\textbf{Asynchronous Override via \texttt{<async>}:}
Concurrency can be implemented in Sync-First mode by defining a function $F_{\text{async}}$ with the corresponding \texttt{<async>\ldots</async>} \textit{function tokens}.
The XML element $E_{\text{async}}$ conforms to:
\begin{equation}
	E_{\text{async}} ::= \text{AFToken}_{\text{async}} \cdot \text{Elements}_{\text{async}}
	\cdot \text{RFToken}_{\text{async}}
\end{equation}
where $\text{Elements}_{\text{async}}=[E_1, E_2, \dots, E_k]$ denotes the sub-elements nested within $E_{\text{async}}$.
The lifecycle $\mathcal{L}(E_{\text{async}}) \to F_{\text{async}}$ reassigns each child element $E_i$ to its module-specific sub-channel, enabling the enclosed functions to execute concurrently:
\begin{equation}
	\forall E_i \in \text{Elements}_{\text{async}}: \alpha(F_i) = \mu(F_i)
\end{equation}
The concurrency boundary is confined to the interior of $E_{\text{async}}$.
The element $F_{\text{async}}$ itself remains on the main channel $C_0$, so it is serialized with all functions before and after it according to Eq.~\ref{eq:main_channel_blocking}.
Specifically, $F_{\text{async}}$ does not begin until all preceding functions on $C_0$ have completed, and all subsequent functions are blocked until every child $\mathcal{L}(E_i)$ within $E_{\text{async}}$ has finished:
\begin{equation}
	\alpha(F_{\text{async}}) = C_0 \;\Rightarrow\; \forall F_j,\ j > i_{\text{async}}: \Gamma(F_{\text{async}}, F_j)
\end{equation}

\subsubsection{Mode Comparison}

The two modes are duals of each other. Async-First defaults to concurrent execution and uses \texttt{<wait>} to enforce sequential ordering, while Sync-First defaults to sequential execution and uses \texttt{<async>} to enable concurrency.
Both modes share the same underlying channel architecture and scheduling laws and differ only in the default channel assignment policy $\alpha$.
The choice between modes depends on the expected task distribution. Async-First is preferable when most behaviors are naturally concurrent (\eg, speaking while gesturing), whereas Sync-First suits scenarios where strict sequential ordering is dominant.

\subsubsection{Text as First-Class Citizen}

Orthogonal to the two scheduling modes above, the scheduler supports a special policy in which textual output is treated as a blocking operation.
When enabled, functions mapped from textual \textit{function tokens} (FChar and FRef), which typically correspond to text-to-speech output, are unconditionally assigned to the main channel $C_0$, thereby blocking all subsequent function calls until the entire text has been processed.
Let $\mathcal{T}\subseteq \mathcal{F}$ denote the subset of such text-derived functions. The constraint is:
\begin{equation}
	\forall\ F_i \in \mathcal{T} \;\Rightarrow\; \alpha(F_i) = C_0
\end{equation}
This policy is not used in our experiments, but is useful in scenarios where speech output must strictly precede or block concurrent physical actions.

  \section{Experiment and Evaluation Details}\label{appendix:evaluation}

This appendix provides supplementary details for the experimental evaluation presented in Section~\ref{sec:experiment}.
We first describe the computation procedure of the DSBC evaluation metric, then define the behavioral execution patterns underlying our task taxonomy.
We subsequently list all 33 evaluation tasks together with the ground truth \textit{function token} sequences for the 30 grounded HRI tasks, and report per-task DSBC scores across all evaluated LLMs.
We then present the long-horizon multimodal task evaluation, including scoring rubrics, per-trial results, and qualitative error analysis.
Finally, we provide a structural analysis of the limitations of native function call interfaces.

\subsection{DSBC Metric Details}\label{appendix:dsbc}

This subsection provides a step-by-step procedural description of the Directed Structured Behavior Correctness (DSBC) metric defined in Section~\ref{sec:experiment}, complementing the formal definitions therein with an algorithmic perspective.

\subsubsection{Design Rationale}

DSBC evaluates how faithfully a generated XML behavior tree reproduces a ground truth tree by performing a layer-by-layer comparison from the root to the deepest node.
At each depth level, the metric separately assesses two complementary aspects:
(1) \textit{structural fidelity}, which measures whether the correct function tags appear in the correct quantities and positions, and
(2) \textit{parametric accuracy}, which measures whether each function tag carries the correct attributes (\eg, direction, duration, speed).
By averaging scores across all layers, DSBC captures both coarse-grained action selection at the top level and fine-grained sub-action specification in nested structures, making it well-suited for evaluating hierarchical behavioral programs.

\subsubsection{Computation Procedure}

Given a ground truth tree $G$ and a generated tree $T$, the DSBC score is computed as follows:

\begin{enumerate}
    \item \textbf{Tree construction.}
    Parse both $G$ and $T$ as XML element trees.
    Compute the maximum depth $D^* = \max(d_G, d_T)$.
    If a layer exists in one tree but not the other, the missing layer is treated as an empty element sequence.

    \item \textbf{Layer-wise evaluation.}
    For each depth level $l = 0, 1, \dots, D^*{-}1$:
    \begin{enumerate}
        \item Extract the ordered element sequences $L_l^G$ and $L_l^T$ at depth $l$.

        \item \textit{Tag Sequence Score} $S_{tag}(l)$: Combines two sub-scores.
        \textbf{Count accuracy} $C_{cnt}(l)$ measures whether each unique tag type appears the same number of times in both sequences, where any over- or under-generation of a tag type incurs a proportional penalty.
        \textbf{Order accuracy} $C_{ord}(l)$ performs strict positional comparison, in which only element pairs at identical indices with matching tag names contribute to the score.
        The tag score is their average: $S_{tag}(l) = (C_{cnt}(l) + C_{ord}(l))/2$.

        \item \textit{Attribute Score} $S_{attr}(l)$: For each ground truth element $e^G$, identifies all generated elements sharing the same tag name as match candidates.
        Among these candidates, computes the fraction of $e^G$'s attribute name-value pairs that are exactly reproduced, and selects the maximum across all candidates.
        The attribute score is the average of these best-match fractions over all ground truth elements at depth $l$.

        \item The layer score combines both aspects equally: $S_l = (S_{tag}(l) + S_{attr}(l))/2$.
    \end{enumerate}

    \item \textbf{Aggregation.}
    The final DSBC score is the arithmetic mean over all $D^*$ layers:
    \begin{equation}
        \text{DSBC}(G, T) = \frac{1}{D^*}\sum_{l=0}^{D^*-1} S_l
    \end{equation}
\end{enumerate}

The resulting score lies in $[0, 1]$, where $1$ indicates perfect structural and parametric correspondence.
This layer-wise decomposition ensures that errors at different depths of the behavioral hierarchy, from top-level action sequencing to nested sub-action parameterization, are weighted equally in the final score.

\subsection{Behavioral Execution Patterns}\label{appendix:behavioral-patterns}

\begin{figure*}[ht]
\centering
\resizebox{0.85\textwidth}{!}{%
\begin{tabular}{@{}c@{\hspace{1em}}c@{}}
	\subfigure[Sequential Execution]{
		\label{fig:sequential_execution}
		\begin{tikzpicture}[scale=1]
			\draw[->, thick] (0,0) -- (6,0) node[right] {\scriptsize Time};
			\node[left] at (0,0) {\scriptsize $T$};
			\foreach \x in {0,...,5} {
					\draw (\x,0.1) -- (\x,-0.1) node[below] {\tiny \x};
				}
			\draw[fill=red!20, draw=black] (0,1.8) rectangle (2,2.3) node[pos=.5] {\scriptsize $F_1$};
			\draw[fill=red!20, draw=black] (2,1.8) rectangle (3.5,2.3) node[pos=.5] {\scriptsize $F_2$};
			\draw[fill=red!20, draw=black] (3.5,1.8) rectangle (5,2.3) node[pos=.5] {\scriptsize $F_3$};
			\node[left] at (0.1,2.05) {\scriptsize $C_1$};
		\end{tikzpicture}
	}
	&
	\subfigure[Parallel Execution]{
		\label{fig:parallel_execution}
		\begin{tikzpicture}[scale=1]
			\draw[->, thick] (0,0) -- (6,0) node[right] {\scriptsize Time};
			\node[left] at (0,0) {\scriptsize $T$};
			\foreach \x in {0,...,5} {
					\draw (\x,0.1) -- (\x,-0.1) node[below] {\tiny \x};
				}
			\draw[fill=red!20, draw=black] (1,1.8) rectangle (5,2.3) node[pos=.5] {\scriptsize $F_1$};
			\node[left] at (0.1,2.05) {\scriptsize $C_1$};
			\draw[fill=yellow!20, draw=black] (0.5,1.2) rectangle (5.5,1.7) node[pos=.5] {\scriptsize $F_2$};
			\node[left] at (0.1,1.45) {\scriptsize $C_2$};
			\draw[fill=blue!20, draw=black] (2,0.6) rectangle (5,1.1) node[pos=.5] {\scriptsize $F_3$};
			\node[left] at (0.1,0.85) {\scriptsize $C_3$};
		\end{tikzpicture}
	}
	\\
	\subfigure[Lifecycle-Scoped Termination]{
		\label{fig:content_driven_termination}
		\begin{tikzpicture}[scale=1]
			\draw[->, thick] (0,0) -- (6,0) node[right] {\scriptsize Time};
			\node[left] at (0,0) {\scriptsize $T$};
			\foreach \x in {0,...,5} {
					\draw (\x,0.1) -- (\x,-0.1) node[below] {\tiny \x};
				}
			\fill[pattern=north east lines, pattern color=gray!100] (0.5,0.25) rectangle (5.5,2.1);
			\draw[->, thick] (4.5,2.4) -- (4.8,2.05);
			\node[align=center, font=\tiny, above] at (3,2.4) {$F_1$ maintains state until $F_2$, $F_3$, $F_4$ complete};

			\draw[fill=red!20, draw=black] (0.5,1.5) rectangle (5.5,2.0) node[pos=.5] {\scriptsize $F_1$};
			\draw[fill=yellow!20, draw=black] (0.5,0.9) rectangle (3.0,1.4) node[pos=.5] {\scriptsize $F_2$};
			\draw[fill=yellow!20, draw=black] (3.0,0.9) rectangle (4.0,1.4) node[pos=.5] {\scriptsize $F_4$};
			\draw[fill=blue!20, draw=black] (1.5,0.3) rectangle (5.5,0.8) node[pos=.5] {\scriptsize $F_3$};
			\node[left] at (0.1,1.75) {\scriptsize $C_1$};
			\node[left] at (0.1,1.15) {\scriptsize $C_2$};
			\node[left] at (0.1,0.55) {\scriptsize $C_3$};
		\end{tikzpicture}
	}
	&
	\subfigure[External-Event Interruption]{
		\label{fig:event_driven_termination}
		\begin{tikzpicture}[scale=1]
			\draw[->, thick] (0,0) -- (6,0) node[right] {\scriptsize Time};
			\node[left] at (0,0) {\scriptsize $T$};
			\foreach \x in {0,...,5} {
					\draw (\x,0.1) -- (\x,-0.1) node[below] {\tiny \x};
				}
			\draw[fill=red!20, draw=black] (0,1.5) rectangle (5.5,2.0) node[pos=.5] {\scriptsize $F_1$};
			\draw[dashed, thick, red] (5.5,2.0) circle(0.1);
			\draw[fill=yellow!20, draw=black] (0.5,0.9) rectangle (5.5,1.4) node[pos=.5] {\scriptsize $F_2$};
			\draw[dashed, thick, red] (5.5,1.4) circle(0.1);
			\draw[->, thick, red] (4,2.3) -- (5.35,2.0);
			\draw[->, thick, red] (4,2.3) -- (5.35,1.4);
			\node[font=\tiny, align=center] at (4,2.4) {external interrupt};
			\node[left] at (0.1,1.75) {\scriptsize $C_1$};
			\node[left] at (0.1,1.15) {\scriptsize $C_2$};
		\end{tikzpicture}
	}
\end{tabular}%
}

	\caption{Behavioral execution patterns. $F$ denotes action functions, and $C$ denotes independent actuator channels.}
	\label{fig:behavioral_patterns}
\end{figure*}

Based on the multi-channel scheduling model defined in Section~\ref{appendix:scheduler}, we identify four fundamental behavioral execution patterns that characterize how function calls are coordinated across channels and over time.
These patterns underlie the five behavioral modes used in our evaluation task design (Section~\ref{appendix:task-descriptions}) and are illustrated in Fig.~\ref{fig:sequential_execution}--\ref{fig:event_driven_termination}.

\textbf{I. Finite Execution Patterns}

\textit{Sequential Execution.}
Actions execute in strict temporal order on a single channel, where each action must complete before the next begins.
As shown in Fig.~\ref{fig:sequential_execution}, actions $F_1$, $F_2$, and $F_3$ execute sequentially on channel $C_1$, with $F_1 \in T[0,2]$, followed by $F_2 \in T[2,3.5]$, and finally $F_3 \in T[3.5,5]$.

\textit{Parallel Execution.}
Multiple actions execute concurrently across independent channels with overlapping temporal intervals.
Fig.~\ref{fig:parallel_execution} demonstrates concurrent execution where $F_1 \in T[1,5]$ on $C_1$, $F_2 \in T[0.5,5.5]$ on $C_2$, and $F_3 \in T[2,5]$ on $C_3$, all with overlapping temporal domains.

\textbf{II. Persistent State Patterns}

\textit{Lifecycle-Scoped Termination.}
A primary action maintains its state until all dependent sub-actions reach completion based on internal state monitoring.
Fig.~\ref{fig:content_driven_termination} shows $F_1 \in T[0.5,5.5]$ on $C_1$ persisting while monitoring the completion of $F_2 \in T[0.5,3]$ and $F_4 \in T[3,4]$ on $C_2$, and $F_3 \in T[1.5,5.5]$ on $C_3$, terminating only when all sub-tasks finish.

\textit{External-Event Interruption.}
Continuous actions terminate abruptly in response to external interrupts, regardless of their internal completion status.
As illustrated in Fig.~\ref{fig:event_driven_termination}, both $F_1 \in T[0,5.5]$ on $C_1$ and $F_2 \in T[0.5,5.5]$ on $C_2$ are terminated at $T=5.5$ by an external interrupt event.

\subsection{Task Instructions}\label{appendix:task-descriptions}

Based on the behavioral execution patterns defined above, we design 30 grounded HRI tasks organized into five behavioral modes of increasing complexity: atomic, sequential, parallel, sequential-parallel composite, and lifecycle-scoped termination.
We further include three long-horizon multimodal tasks that require multi-turn reasoning with visual observations.
Each task is expressed as a natural language instruction issued to the robot \rbtname.
The complete task list is provided in Table~\ref{tab:task-descriptions}.

\subsection{Ground Truth Function Tokens}\label{appendix:ground-truth}

Table~\ref{tab:ground-truth} presents the ground truth \textit{function token} sequences for the 30 grounded HRI tasks (Tasks 1--30).
Each entry shows the expected XML output that the LLM should generate in response to the corresponding natural language instruction in Table~\ref{tab:task-descriptions}.
Plain text outside XML tags represents FChar tokens (\eg, speech output), while \texttt{<wait>} elements enforce sequential ordering via the main channel as described in Section~\ref{appendix:scheduler}.
These sequences serve as the reference for computing the DSBC scores reported in Section~\ref{appendix:detailed-scores}.

\subsection{Per-Task Performance}\label{appendix:detailed-scores}

Table~\ref{tab:task_scores} reports the per-task DSBC scores for the 30 grounded HRI tasks across all 21 evaluated LLMs.
For each model-task combination, we conduct five independent rounds and report the mean score to mitigate variance from model nondeterminism.
Task categories in Table~\ref{tab:task_scores} correspond to the groupings defined in Table~\ref{tab:task-descriptions}.
The micro-average (mean over all cases per round) and macro-average (mean of per-category means per round) rows report the mean$\pm$std across rounds.
We highlight several patterns that emerge from the per-task breakdown and complement the category-level analysis in Section~\ref{sec:experiment}.

\subsubsection{Universally Easy \vs Hard Tasks}
Three tasks achieve near-perfect scores across all 21 models: Task~3 (head nod), Task~11 (four-leg sequential tapping), and Task~13 (shake butt then creep backward), all scoring $\geq$0.89.
These tasks share a common trait: their function tokens involve unambiguous, single-tag action names with simple or no attribute sets, allowing even lower-performing models to generate correct XML reliably.

In contrast, four tasks prove universally challenging.
Task~9 (walk in a square with 20\,cm sides) yields the lowest scores across the board, with no model exceeding 0.46 (DeepSeek-V3.2).
This task requires decomposing a geometric trajectory into a four-iteration loop of \texttt{run\_distance} and \texttt{rotate} calls with precise distance and angle parameters, a form of spatial planning that current LLMs struggle to express as structured function tokens.
Task~5 (nod then wave) produces scores of exactly 0.35 for 14 of 21 models, indicating systematic failure to generate the correct \texttt{<wave/>} tag; only Qwen3-235B achieves 1.00, while Claude-Sonnet-4.5 (0.87) and DeepSeek-V3.1 (0.87) attain partial matches.
Tasks~10 and~12 also exhibit predominantly low scores ($\leq$0.35), where the combination of duration-based parameterization and sequential \texttt{<wait>} wrapping consistently challenges models.

\subsubsection{Duration and Numerical Attributes}
Tasks involving explicit duration or angle specifications reveal a systematic attribution weakness.
Task~16 (stand backward while nodding for 3\,s) and Task~14 (smile and say hello simultaneously) are both parallel tasks, differing primarily in the requirement for a \texttt{duration} attribute.
Task~14 achieves scores $\geq$0.75 for all models, whereas Task~16 exhibits a bimodal distribution: 11 models score exactly 0.50 (correct tags, missing duration) while 6 models achieve 1.00 (full attribute specification).
This bimodality indicates that for parallel tasks, attribute generation rather than structural topology is the primary performance bottleneck.
A similar pattern appears in Task~17 (frustrated expression while flapping ears for 6\,s), where scores range from 0.25 to 1.00 depending on whether models correctly parameterize the duration.

\subsubsection{Sequential Complexity Scaling}
Within the sequential category, performance degrades as the number of sequential steps and action heterogeneity increase.
Task~4 (two-step: say hello $\to$ smile) and Task~6 (two-step with specialized motor actions) average 0.94 and 0.93 across models, respectively.
Task~11 (four-step leg tapping) similarly achieves near-perfect scores despite having four sequential elements.
However, Task~8 (five-step: greet $\to$ shake ears $\to$ shake head $\to$ sit $\to$ bark) drops sharply, with 12 of 21 models scoring below 0.56.
The contrast between Task~11 and Task~8 is informative: Task~11 involves structurally identical repeated elements differing only in leg index and tap count, while Task~8 requires diverse action types interleaved with speech.
This suggests that \textit{heterogeneous} sequential composition amplifies generation errors substantially beyond what step count alone would predict.

\subsubsection{Lifecycle-Scoped Termination as a Model Discriminator}
The lifecycle-scoped termination category (Tasks~23--30) exhibits the highest inter-model variance and serves as the strongest discriminator of model capability.
Task~23 (nod until 3 steps backward complete) is particularly revealing: 8 of 21 models score exactly 0.00, indicating complete failure to produce the nested XML structure, while GPT-4.1 and Grok-3 achieve perfect 1.00.
This binary outcome suggests that the concept of parent-action persistence conditioned on child-action completion is either wholly absent or reliably present in a model's generative repertoire, with little gradation.
Tasks~29 and~30 test whether models correctly sequence sub-actions within an identical persistent parent (\texttt{<shake\_like\_dog>}); scores range from 0.34 to 1.00, and notably, swapping the order of sub-tasks between these two otherwise identical tasks produces only minor score variations within the same model, suggesting that once a model grasps the nesting concept, internal reordering is handled robustly.

\subsubsection{Model-Specific Per-Task Observations}
Several per-task results reveal model-specific characteristics not visible from category-level averages.
Claude-Sonnet-4 is the only model to score 1.00 on Task~12 (turn head 90° right then rotate body 180° left), a task where all other models score $\leq$0.35.
This indicates superior instruction-following for precise numerical parameter specifications involving spatial coordinates.
Conversely, Qwen-Plus achieves 0.00 on Task~1 (right-eye wink), the simplest task in the evaluation, due to failure to generate the correct emotion identifier \texttt{wink\_right}. This is an unexpected outlier given its otherwise competitive performance on sequential-parallel composite tasks (category average 0.76).
DeepSeek-V3.1 is the only model family to score below 1.00 on all three atomic tasks (0.85, 0.84, 1.00), suggesting that its function tag generation does not fully align with the provided API definitions even for single-action commands.
Among the Qwen model family, large performance gaps exist between versions: Qwen-Flash (micro-avg 0.58) and Qwen-Turbo (0.60) substantially underperform Qwen3-Max (0.72) and Qwen-Plus (0.66), with Task~7 exemplifying the disparity (Qwen-Flash: 0.07 \vs Qwen3-Max: 0.77).

\subsubsection{Newer Models Do Not Guarantee Higher Scores}
A counter-intuitive but consistent pattern emerges across multiple model families: successor models that claim improvements over their predecessors on general benchmarks do not reliably outperform them on our structured behavioral generation tasks.

The most striking example is the Claude family.
Claude-Sonnet-4 achieves the highest overall micro-average (0.83), but its successor Claude-Sonnet-4.5 drops to 0.73, representing a 10-point regression.
The degradation is particularly severe in the lifecycle-scoped termination category, where Claude-Sonnet-4.5 scores 0.00 on Task~23 (\vs 0.80 for Claude-Sonnet-4) and 0.45 on Task~24 (\vs 1.00).
Meanwhile, the older Claude-Sonnet-3.7 (micro-avg 0.77) outperforms its newer sibling Claude-Sonnet-4.5 as well.

A similar pattern holds across OpenAI models.
GPT-5.2, despite being a generation ahead, achieves the same micro-average as GPT-4.1 (both 0.74), and GPT-4o scores a comparable 0.73.
Across three successive model releases, no measurable improvement is observed on our benchmark.

The trend extends to Chinese model families.
Doubao-Seed-1.8 (micro-avg 0.66) underperforms its predecessor Doubao-Seed-1.6 (0.71), with notable drops on tasks such as Task~25 (0.68 \vs 1.00), Task~21 (0.49 \vs 1.00), and Task~29 (0.39 \vs 0.81).
Similarly, DeepSeek-V3.2 (0.65) scores lower than DeepSeek-V3.1 (0.67), despite being a more recent release.

These results suggest that the capabilities tested by our benchmark, namely generating well-formed hierarchical XML structures with correct nesting, topological ordering, and precise parametric attributes, are not well-correlated with the general reasoning and instruction-following abilities that drive improvements on conventional benchmarks.
Structured behavioral generation appears to rely on a distinct skill set, one that may even be degraded by training procedures that prioritize conversational fluency or safety alignment at the expense of precise structural output.

\subsection{Long-Horizon Task Scoring Rubrics}\label{appendix:long-horizon-scoring}

Unlike the grounded HRI tasks evaluated with DSBC, the three long-horizon multimodal tasks (Tasks 31--33) involve open-ended multi-turn reasoning where ground truth token sequences cannot be predefined.
We therefore adopt a human evaluation protocol in which each task is scored on a 10-point scale.
Scores are assigned based on task-specific sub-criteria that assess the key competencies required for successful completion.
The rubrics for each task are detailed below.

\subsubsection{Bottle Collision Detection and Manipulation (10 pts)}
\begin{itemize}
    \item \textbf{Pre-collision observation} (2 pts): The robot observes and assesses the environment before attempting to strike.
    \item \textbf{Appropriate collision action} (2 pts): The robot selects and executes a suitable action to knock over the bottle.
    \item \textbf{Task completion within 5 rounds} (2 pts): The robot successfully knocks over the bottle within five interaction rounds.
    \item \textbf{Outcome judgment accuracy} (4 pts): The robot correctly determines whether the bottle has been knocked over after each attempt.
\end{itemize}

\subsubsection{Rock-Paper-Scissors Game (10 pts)}
\begin{itemize}
    \item \textbf{Immediate observation after selection} (1 pt): The robot captures a visual observation immediately after presenting its own choice.
    \item \textbf{Gesture recognition accuracy} (4 pts): The robot correctly identifies the human player's hand gesture (rock, paper, or scissors).
    \item \textbf{Win/loss logic correctness} (5 pts): The robot correctly applies the game rules to determine the round outcome.
\end{itemize}

\subsubsection{Mirror-Based Self-Recognition (10 pts)}
\begin{itemize}
    \item \textbf{Embodied behavior consistency} (1 pt): The robot's responses are consistent with its embodied agent identity.
    \item \textbf{Active motion-based verification} (4 pts): The robot performs deliberate actions (\eg, waving, nodding) and observes mirror reflections to test whether the reflected entity mirrors its own movements.
    \item \textbf{Conclusion correctness} (3 pts, partial credit):
    \begin{itemize}
        \item 3 pts: Actions are clearly distinguishable and the robot draws a correct conclusion, ruling out visual hallucination.
        \item 1 pt: The robot cannot determine the result with certainty.
        \item 0 pts: Actions fail to produce observable mirror feedback.
    \end{itemize}
    \item \textbf{Repeated verification} (2 pts): The robot repeats the experiment to confirm or revise its initial conclusion.
\end{itemize}

\subsection{Long-Horizon Per-Trial Scores}\label{appendix:long-horizon-scores}

Table~\ref{tab:long-horizon-per-trial} reports the individual trial scores for all three long-horizon multimodal tasks across the three evaluated LMMs.
Each task is conducted three independent times per model, and the averages of these trials correspond to the summary scores reported in Table~\ref{tab:long_horizon}.
Notable variance is observed in the Bottle Collision and Mirror Self-Recognition tasks, where trial outcomes depend heavily on stochastic factors such as initial positioning and visual interpretation.
In contrast, the RPS Game exhibits low variance, indicating that gesture recognition and rule application are relatively stable across runs.
We discuss each task in detail below.

\textbf{Bottle Collision.}
GPT-4.1 achieves the highest average (8.7) with per-trial scores of 10.0, 6.0, and 10.0.
The single mid-range trial (T2 = 6.0) reflects misjudgment of the collision outcome rather than locomotion failure, as the model reached the bottle but incorrectly assessed whether it had been toppled.
Claude-Sonnet-4 shows an inverted pattern (4.0, 10.0, 10.0), where the initial trial suffers from insufficient approach distance, a failure that is corrected in subsequent trials through more aggressive locomotion parameters.
Doubao-Seed-1.6 produces the most consistent but lowest scores (6.0, 6.0, 4.0), with \textit{frequent} insufficient walking distance (Table~\ref{tab:long-horizon-errors}) indicating that this model's motion planning is systematically conservative regardless of task feedback.

\textbf{Rock-Paper-Scissors.}
All three models achieve high and stable scores (9.0--10.0 per trial), with the total range spanning only 1 point.
The near-ceiling performance confirms that the RPS task primarily tests visual recognition and deterministic rule application, both of which are robust and well-calibrated competencies in current LMMs.
The single 10.0 trials (Doubao T3 and Claude T1) represent runs with perfect gesture identification and immediate visual observation, while 9.0 trials typically lose 1 point for delayed visual capture after the model's own choice declaration, a procedural error noted for all three models in Table~\ref{tab:long-horizon-errors}.

\textbf{Mirror Self-Recognition.}
The mirror test reveals the most striking variance pattern.
Both Claude-Sonnet-4 and GPT-4.1 achieve 8.0 on their first trial but collapse to 1.0 on both subsequent trials, yielding averages of 3.3 despite demonstrating task competence in T1.
Doubao-Seed-1.6 follows a similar but weaker trajectory (3.0, 1.0, 1.0; average 1.7).
A score of 8.0 aligns with successfully performing active verification and reaching a correct conclusion while missing the repeated verification criterion (2 pts); a score of 1.0 indicates that only the embodied behavior consistency point (1 pt) was awarded, with all substantive verification and reasoning components failing entirely.
The T1 successes suggest that models \textit{can} execute the deliberate action--observation--conclusion loop required for mirror self-identification when stochastic sampling produces an appropriate reasoning chain.
However, the T2/T3 collapses indicate that this capability is fragile: GPT-4.1 \textit{frequently} skips verification and guesses conclusions without performing actions, while Claude-Sonnet-4 \textit{frequently} performs only a single verification without repeating to confirm (Table~\ref{tab:long-horizon-errors}).
These patterns suggest that embodied self-recognition through mirror interaction is not a stable capability of current LMMs but rather emerges sporadically, and that the underlying causal reasoning, which connects self-initiated motor commands with their visual consequences, is easily bypassed by the model's tendency toward premature conclusion.

\begin{table}[htbp]
\centering
\caption{Per-Trial Scores for Long-Horizon Multimodal Tasks}
\label{tab:long-horizon-per-trial}
\renewcommand{\arraystretch}{1.15}
\setlength{\tabcolsep}{3pt}
\scriptsize
\begin{tabular}{@{} l l c c c c @{}}
    \toprule
    \textbf{Task} & \textbf{Model} & \textbf{T1} & \textbf{T2} & \textbf{T3} & \textbf{Avg.} \\
    \midrule
    \multirow{3}{*}{\shortstack[l]{Bottle Collision}}
    & Doubao  & 6.0 & 6.0 & 4.0 & 5.3 \\
    & Claude  & 4.0 & 10.0 & 10.0 & 8.0 \\
    & GPT-4.1 & 10.0 & 6.0 & 10.0 & 8.7 \\
    \midrule
    \multirow{3}{*}{\shortstack[l]{RPS Game}}
    & Doubao  & 9.0 & 9.0 & 10.0 & 9.3 \\
    & Claude  & 10.0 & 9.0 & 9.0 & 9.3 \\
    & GPT-4.1 & 9.0 & 9.0 & 9.0 & 9.0 \\
    \midrule
    \multirow{3}{*}{\shortstack[l]{Mirror Test}}
    & Doubao  & 3.0 & 1.0 & 1.0 & 1.7 \\
    & Claude  & 8.0 & 1.0 & 1.0 & 3.3 \\
    & GPT-4.1 & 8.0 & 1.0 & 1.0 & 3.3 \\
    \bottomrule
\end{tabular}
\end{table}

\subsection{Qualitative Error Analysis}\label{appendix:error-analysis}

Table~\ref{tab:long-horizon-errors} summarizes the primary failure modes observed across the three long-horizon tasks during human evaluation.
In the table, \textbf{frequent} denotes $\geq$3 occurrences, \textit{occasional} denotes 1--2 occurrences, and -- indicates the issue was not observed.
We identify three recurring issues across models: (1) insufficient locomotion distance causing the robot to miss physical targets, (2) responses inconsistent with visual observations due to recognition errors or hallucination, and (3) miscellaneous procedural errors.

\begin{table*}[ht]
\centering
\caption{Observed failure modes in long-horizon tasks.}
\label{tab:long-horizon-errors}
\renewcommand{\arraystretch}{1.2}
\footnotesize
\begin{tabular}{@{} l l c c c @{}}
    \toprule
    \textbf{Task} & \textbf{Failure Mode} & \textbf{Doubao} & \textbf{Claude} & \textbf{GPT-4.1} \\
    \midrule
    \multirow{2}{*}{Bottle Knock-over}
    & Insufficient walking distance, failed to reach the bottle & \textbf{frequent} & -- & \textit{occasional} \\
    & Hallucinated success, claimed bottle was knocked over without contact & \textit{occasional} & \textit{occasional} & -- \\
    \midrule
    \multirow{2}{*}{Rock-Paper-Scissors}
    & Failed to invoke visual observation immediately after making a choice & \textbf{frequent} & \textbf{frequent} & \textbf{frequent} \\
    & Incorrect win/loss judgment & \textit{occasional} & -- & -- \\
    \midrule
    \multirow{3}{*}{Mirror Self-Recognition}
    & Skipped verification, guessed conclusion without performing actions & \textit{occasional} & -- & \textbf{frequent} \\
    & Performed only a single verification without repetition & -- & \textbf{frequent} & \textit{occasional} \\
    & Drew conclusion without observing action outcome (hallucination) & -- & \textit{occasional} & -- \\
    \bottomrule
\end{tabular}
\end{table*}

The error patterns reveal distinct model-specific failure strategies across the three tasks.

\textbf{Locomotion and spatial planning.}
The bottle collision task exposes limitations in spatial reasoning during multi-turn physical interaction.
Doubao-Seed-1.6 \textit{frequently} underestimates the required approach distance, generating locomotion commands with insufficient step counts or stride lengths, leading to repeated failed attempts that exhaust the five-round interaction budget without physical contact.
GPT-4.1 exhibits this failure only \textit{occasionally}, indicating better calibration of locomotion parameters, consistent with its higher average score (8.7 \vs 5.3).
Claude-Sonnet-4 avoids this failure mode entirely, suggesting that its action parameterization is better grounded in the physical task requirements through more effective use of visual observations to adjust locomotion.

\textbf{Visual verification and hallucination.}
Two qualitatively distinct failure modes involve visual observations.
In the bottle task, models \textit{occasionally} hallucinate success by claiming the bottle has been knocked over when no physical contact has occurred, representing a reasoning-observation disconnect where the model's expectations override actual visual input.
In the RPS task, all three models \textit{frequently} fail to invoke visual observation \textit{immediately} after declaring their own choice, instead relying on observations captured before or during the declaration.
Although this procedural error does not always yield incorrect outcomes (gesture recognition remains accurate when observation occurs), it introduces temporal misalignment between the model's action and its perceptual input that could be consequential in more dynamic settings.

\textbf{Embodied self-recognition as a capability frontier.}
The mirror test failure modes form a hierarchical spectrum of reasoning depth.
At the shallowest level, models skip verification entirely and guess a conclusion (GPT-4.1: \textit{frequent}).
At an intermediate level, models perform a single verification without repeating to confirm (Claude: \textit{frequent}).
At the deepest failure mode, models draw conclusions before observing the outcome of their own actions, constituting a causal reasoning hallucination (Claude: \textit{occasional}).
The predominance of shallow failure modes (skipping or single verification) over active hallucination suggests that the primary bottleneck is not visual processing but the models' inability to maintain an agentic verification loop that iteratively acts, observes, and revises conclusions, which is a prerequisite for embodied self-recognition through mirror interaction.
Notably, Doubao-Seed-1.6 is the only model observed to \textit{occasionally} skip verification, yet its scores are the lowest (average 1.7), indicating that once the active verification loop fails, no compensatory reasoning pathway exists.

\subsection{Structural Limitations of Native Function Calls}\label{appendix:funcall-analysis}

This subsection provides empirical evidence and structural analysis of the limitations of native function call interfaces (\eg, OpenAI tool calls), complementing the summary presented in Table~\ref{tab:function_call}.

\subsubsection{Empirical Results}

Table~\ref{tab:funcall-detail} reports the actual \texttt{tool\_calls} output produced by GPT-4.1 for each task in Table~\ref{tab:function_call}, alongside the expected behavior.
We observe three recurring failure modes:
\begin{enumerate}
    \item \textbf{Linguistic-behavioral output decoupling.}
    Tasks requiring simultaneous or interleaved speech and motor actions (Tasks~6, 9, 11, 12, 14) consistently fail to produce any speech output.
    In Task~6, the model emits only a single \texttt{spiderMove} call, omitting both ``say hello'' and ``say goodbye'' entirely, while the \texttt{content} field remains empty.
    In Task~9, the speech is similarly omitted, and the model further hallucinates an unrequested \texttt{smile} action, substituting a facial expression for the missing verbal output.
    In Tasks~11 and~12, all speech acts (``stepping back'' and ``say hello/goodbye'', respectively) are lost alongside substantial action omission (see below).
    In Task~14, the greeting speech is entirely absent despite being the child action whose completion should terminate the parent.

    \item \textbf{Action omission under structural complexity.}
    As task structure exceeds simple flat parallelism or two-step sequences, native function calls progressively drop actions.
    Task~4 (two-step sequence) succeeds, but Task~5 (three-step: forward $\to$ rotate $\to$ forward) produces only a single \texttt{spiderMove}, with both the rotation and the second forward movement lost.
    The sequential-parallel composite tasks exhibit the most severe omission:
    Task~10 (forward $\to$ rotate+shake) retains only \texttt{spiderMove} (1 of 3 expected actions);
    Task~11 (smile+ears $\to$ backward+say) retains only \texttt{smile} and \texttt{shakeEars} from the first parallel group (2 of 4), with the entire second phase lost;
    Task~12 (say $\to$ rotate+shake $\to$ say) retains only \texttt{spiderRotate} (1 of 4), losing both speech acts and the parallel head shake.

    \item \textbf{Structural inexpressibility.}
    Tasks~13--15 (Lifecycle-Scoped Termination) require a parent action to persist while monitoring the completion of nested child actions, forming a tree-structured execution lifecycle that is fundamentally beyond the representational capacity of the flat JSON array format.
\end{enumerate}

\begin{table*}[ht]
\centering
\caption{Native Function Call Outputs (GPT-4.1) vs.\ Expected Behavior}
\label{tab:funcall-detail}
\renewcommand{\arraystretch}{1.2}
\footnotesize
\begin{tabular}{@{} c p{6cm} p{8.5cm} @{}}
    \toprule
    \textbf{\#} & \textbf{Expected Actions} & \textbf{Actual \texttt{tool\_calls}} \\
    \midrule
    1 & say & \texttt{(content: ``Hello!'')} \cmark \\
    2 & headAct(nod, 3s) & \texttt{headAct(nod, 3)} \cmark \\
    3 & pushUp(3) & \texttt{pushUp(3)} \cmark \\
    \midrule
    4 & move $\to$ headShake & \texttt{spiderMove}, \texttt{headAct(shake)} \cmark \\
    5 & move $\to$ rotate $\to$ move & \texttt{spiderMove} \xmark{\scriptsize~(2 actions lost)} \\
    6 & say $\to$ move $\to$ say & \texttt{spiderMove} \xmark{\scriptsize~(all speech lost)} \\
    \midrule
    7 & move \textbar\textbar{} headShake & \texttt{spiderMove}, \texttt{headAct(shake)} \cmark \\
    8 & move \textbar\textbar{} headShake \textbar\textbar{} earShake & \texttt{spiderMove}, \texttt{headAct(shake)}, \texttt{shakeEars} \cmark \\
    9 & move \textbar\textbar{} say & \texttt{spiderMove}, \texttt{smile} \xmark{\scriptsize~(speech lost, hallucinated)} \\
    \midrule
    10 & move $\to$ (rotate \textbar\textbar{} headShake) & \texttt{spiderMove} \xmark{\scriptsize~(2 actions lost)} \\
    11 & (smile \textbar\textbar{} ears) $\to$ (backward \textbar\textbar{} say) & \texttt{smile}, \texttt{shakeEars} \xmark{\scriptsize~(2nd phase lost, speech lost)} \\
    12 & say $\to$ (rotate \textbar\textbar{} headShake) $\to$ say & \texttt{spiderRotate} \xmark{\scriptsize~(3 actions lost, all speech lost)} \\
    \midrule
    13--15 & nested lifecycle & \textit{structurally inexpressible} \xmark \\
    \bottomrule
\end{tabular}
\end{table*}

\subsubsection{Structural Analysis}

The above failures stem from two fundamental design constraints of the native function call protocol.

The most critical limitation for embodied agent control is the \textit{architectural separation of linguistic and executable outputs}.
In the API response schema, natural language text and structured tool invocations are routed through two disjoint output fields (\texttt{content} and \texttt{tool\_calls}).
Although both fields may technically co-occur within a single response, current LLMs overwhelmingly populate only the \texttt{tool\_calls} field when tool definitions are provided, effectively suppressing all free-form text generation.
This renders speech acts, a fundamental component of social human-robot interaction, systematically unreachable through the function call pathway.
Crucially, this is not an implementation artifact but a deliberate protocol design choice that treats text and tool invocations as separate response types rather than interleaved elements of a unified behavioral program.
This separation served as a primary motivation for our development of the \textit{function token} protocol, which unifies speech and action within a single generative token stream.

The second limitation is the flat, topology-free structure of the \texttt{tool\_calls} array.
Each invocation is an independent JSON object with no mechanism to encode execution ordering, concurrency scoping, or parent-child lifecycle dependencies.
Even if integrated with a multi-channel scheduler, the scheduler cannot derive the intended execution topology from a flat call list.
In contrast, our XML-based \textit{function tokens} encode topology explicitly: the \texttt{<wait>} element delimits sequential phases, sibling elements within the same scope execute in parallel, and parent elements persist until all children complete.
This structural expressiveness enables \textit{function tokens} to represent all five behavioral modes (Section~\ref{appendix:behavioral-patterns}), whereas native function calls are limited to atomic flat invocations.

\begin{table*}[ht]
\centering
\caption{Task Instructions for Evaluation}
\label{tab:task-descriptions}
\renewcommand{\arraystretch}{1.15}
\footnotesize
\begin{tabular}{@{} >{\bfseries}l p{13.5cm} @{}}
    \toprule
    Task ID & \textbf{Instruction} \\
    \midrule

    \multicolumn{2}{l}{\cellcolor{gray!20}\textbf{Atomic Command}} \\
    Task 1 & Wink with your right eye \\
    Task 2 & Do push-ups \\
    Task 3 & Nod your head \\
    
    \multicolumn{2}{l}{\cellcolor{gray!20}\textbf{Sequential Command}} \\
    Task 4 & First say `hello', then give me a smile \\
    Task 5 & Now first nod your head, then do a wave motion \\
    Task 6 & First do a chest expansion exercise, then do a hand-rubbing motion \\
    Task 7 & Now please shrug your shoulders to the left, then shrug to the right, repeat three times in total, and finally say `I can really dance' \\
    Task 8 & First say `Hello everyone', then shake your ears, then shake your head, then squat like a dog, and finally make a barking sound \\
    Task 9 & Walk in a square with 20cm sides \\
    Task 10 & Wriggle your whole body for 3 seconds, then shake your ears for 2 seconds \\
    Task 11 & Tap the ground with your front right, front left, back left, and back right legs 1, 2, 3, and 4 times respectively \\
    Task 12 & Turn your head 90 degrees to the right, then rotate your whole body 180 degrees to the left \\
    Task 13 & First shake your butt twice at the fastest speed, then creep backward at speed 50 \\
    
    \multicolumn{2}{l}{\cellcolor{gray!20}\textbf{Parallel Command}} \\
    Task 14 & Please smile at me and say `hello' at the same time! \\
    Task 15 & Make a happy expression, nod your head at the same time, also do a push-up, and move your ears too \\
    Task 16 & Stand your body backward while nodding for 3 seconds \\
    Task 17 & Act very frustrated while flapping your ears for 6 seconds \\
    
    \multicolumn{2}{l}{\cellcolor{gray!20}\textbf{Sequential-Parallel Composite Command}} \\
    Task 18 & Spin in place, then play background music while shaking your butt \\
    Task 19 & Shake your ears while playing some background music, then wink both eyes \\
    Task 20 & Perk up your ears, then move forward while playing music, and finally stop to do a push-up \\
    Task 21 & First do 3 seconds push-ups and play music, then happily say `what are you doing' \\
    Task 22 & First play music while standing up, then take two steps forward, and finally play music and sit down \\
    
    \multicolumn{2}{l}{\cellcolor{gray!20}\textbf{Lifecycle-Scoped Termination Command}} \\
    Task 23 & Keep nodding until you finish taking 3 steps backward \\
    Task 24 & Keep shaking your butt while playing music for 10 seconds and showing a calm expression for 3 seconds, stop shaking only after both actions are complete \\
    Task 25 & Maintain music playing and standing posture, finish after nodding continuously for 3 seconds \\
    Task 26 & Execute in order: first shake ears 3 times, then quickly nod three times, then take two steps forward, all accompanied by background music \\
    Task 27 & Shake your body like a dog, then quickly turn your head to the right, then show a surprised expression, only stop shaking after everything is done \\
    Task 28 & Maintain a forward standing posture, then wink your right eye for five seconds while taking a photo, restore your body posture after all actions are completed \\
    Task 29 & Shake like a dog, then shake your ears for 3 seconds, then nod while looking very happy, stop shaking like a dog after all actions are completed \\
    Task 30 & Shake like a dog, then nod while looking very happy, then gently shake your ears for 3 seconds, stop dog shaking after all actions are completed \\
    
    \multicolumn{2}{l}{\cellcolor{gray!20}\textbf{Long-Horizon Multimodal Command}} \\
    Task 31 & Knock down the bottles in front. \\
    Task 32 & Let's play rock-paper-scissors together! You count to three, and when you get to three, say whether you want to play rock, scissors, or paper. Meanwhile, I will show you the gestures for rock-paper-scissors in front of you. Look at my gestures and compare them to what you just chose to see who wins and who loses. If you lose, you have to make a very sad and exaggerated gesture. If you win, you have to make a very happy and exaggerated gesture. \\
    Task 33 & There are two robots in the mirror in front of you and tell me which one is your image. \\
    \bottomrule
\end{tabular}
\end{table*}

\begin{table*}[ht]
\centering
\caption{Ground Truth Function Token Sequences for All Evaluation Tasks}
\label{tab:ground-truth}
\renewcommand{\arraystretch}{1.3}
\footnotesize
\begin{tabular}{@{} >{\bfseries}l p{13.5cm} @{}}
    \toprule
    Task ID & \textbf{Ground Truth (XML Function Tokens)} \\
    \midrule

    \multicolumn{2}{l}{\cellcolor{gray!20}\textbf{Atomic Command}} \\
    Task 1 & \texttt{<face\_emo emo="wink\_right"/>} \\
    Task 2 & \texttt{<push\_up/>} \\
    Task 3 & \texttt{<head\_act action="nod"/>} \\

    \multicolumn{2}{l}{\cellcolor{gray!20}\textbf{Sequential Command}} \\
    Task 4 & \texttt{<wait>Hello</wait>\allowbreak<face\_emo emo="smile"/>} \\
    Task 5 & \texttt{<wait><head\_act action="nod"/></wait>\allowbreak<wave/>} \\
    Task 6 & \texttt{<chest\_expansion\_exercise/>\allowbreak<fly\_rub\_hands/>} \\
    Task 7 & \texttt{<body\_lean direction="left"/>\allowbreak<body\_lean direction="right"/>} $\times 3$ \texttt{<wait>I can really dance!</wait>} \\
    Task 8 & \texttt{<wait>Hello everyone</wait>\allowbreak<wait><shake\_ear/></wait>\allowbreak<wait><head\_act action="shake"/></wait>\allowbreak<wait><sit style="dog"/></wait>\allowbreak{}Woof woof woof} \\
    Task 9 & \texttt{(<run\_distance distance="20"/>\allowbreak<rotate angle="-90"/>)} $\times 4$ \\
    Task 10 & \texttt{<wait><wriggle duration="3"/></wait>\allowbreak<shake\_ear duration="2"/>} \\
    Task 11 & \texttt{<single\_leg\_knock leg="1" times="1"/>\allowbreak<single\_leg\_knock leg="2" times="2"/>\allowbreak<single\_leg\_knock leg="3" times="3"/>\allowbreak<single\_leg\_knock leg="4" times="4"/>} \\
    Task 12 & \texttt{<wait><head\_pose yaw="-90"/></wait>\allowbreak<rotate angle="180"/>} \\
    Task 13 & \texttt{<shake\_ass speed="100" times="2"/>\allowbreak<creep direction="backward" speed="50"/>} \\

    \multicolumn{2}{l}{\cellcolor{gray!20}\textbf{Parallel Command}} \\
    Task 14 & \texttt{Hello\allowbreak<face\_emo emo="smile"/>} \\
    Task 15 & \texttt{<face\_emo emo="happy"/>\allowbreak<head\_act action="nod"/>\allowbreak<push\_up/>\allowbreak<shake\_ear/>} \\
    Task 16 & \texttt{<stand posture="backward"/>\allowbreak<head\_act action="nod" duration="3"/>} \\
    Task 17 & \texttt{<face\_emo emo="very\_frustrated"/>\allowbreak<shake\_ear duration="6"/>} \\

    \multicolumn{2}{l}{\cellcolor{gray!20}\textbf{Sequential-Parallel Composite Command}} \\
    Task 18 & \texttt{<rotate/>\allowbreak<wait>\allowbreak<play\_bgm/>\allowbreak<shake\_ass/>\allowbreak</wait>} \\
    Task 19 & \texttt{<shake\_ear/>\allowbreak<play\_bgm/>\allowbreak<wait>\allowbreak<face\_emo emo="wink"/>\allowbreak</wait>} \\
    Task 20 & \texttt{<ear\_state state="inquisitive"/>\allowbreak<wait><move\_forward/><play\_bgm/></wait>\allowbreak<push\_up/>} \\
    Task 21 & \texttt{<wait><push\_up duration="3"/>\allowbreak<play\_bgm/>\allowbreak</wait>\allowbreak<face\_emo emo="happy"/>\allowbreak{}What are you doing} \\
    Task 22 & \texttt{<wait><play\_bgm/><stand/></wait>\allowbreak<wait><move\_strides direction="forward" times="2"/></wait>\allowbreak<play\_bgm/><sit/>} \\

    \multicolumn{2}{l}{\cellcolor{gray!20}\textbf{Lifecycle-Scoped Termination Command}} \\
    Task 23 & \texttt{<head\_act action="nod">\allowbreak<move\_strides direction="backward" times="3"/>\allowbreak</head\_act>} \\
    Task 24 & \texttt{<shake\_ass>\allowbreak<play\_bgm duration="10"/>\allowbreak<face\_emo emo="calm" duration="3"/>\allowbreak</shake\_ass>} \\
    Task 25 & \texttt{<play\_bgm>\allowbreak<stand>\allowbreak<head\_act action="nod" duration="3"/>\allowbreak</stand>\allowbreak</play\_bgm>} \\
    Task 26 & \texttt{<play\_bgm>\allowbreak<wait>\allowbreak<shake\_ear/>$\times 3$\allowbreak</wait>\allowbreak<wait>\allowbreak<head\_act action="nod"/>$\times 3$\allowbreak</wait>\allowbreak<wait>\allowbreak<move\_strides direction="forward" times="2"/>\allowbreak</wait>\allowbreak</play\_bgm>} \\
    Task 27 & \texttt{<shake\_like\_dog>\allowbreak<wait>\allowbreak<head\_pose yaw="-90"/>\allowbreak</wait>\allowbreak<wait>\allowbreak<face\_emo emo="surprise"/>\allowbreak</wait>\allowbreak</shake\_like\_dog>} \\
    Task 28 & \texttt{<stand posture="forward">\allowbreak<wait>\allowbreak<face\_emo emo="wink\_right" duration="5"/>\allowbreak<capture/>\allowbreak</wait>\allowbreak</stand>} \\
    Task 29 & \texttt{<shake\_like\_dog>\allowbreak<wait>\allowbreak<shake\_ear duration="3"/>\allowbreak</wait>\allowbreak<wait>\allowbreak<head\_act action="nod"/>\allowbreak<face\_emo emo="happy"/>\allowbreak</wait>\allowbreak</shake\_like\_dog>} \\
    Task 30 & \texttt{<shake\_like\_dog>\allowbreak<wait>\allowbreak<head\_act action="nod"/>\allowbreak<face\_emo emo="happy"/>\allowbreak</wait>\allowbreak<wait>\allowbreak<shake\_ear duration="3"/>\allowbreak</wait>\allowbreak</shake\_like\_dog>} \\
    \bottomrule
\end{tabular}
\end{table*}

\begin{sidewaystable*}
    \centering
    \caption{Per-Task DSBC Scores for Grounded HRI Tasks (Tasks 1--30). Micro/macro averages report mean$\pm$std across 5 rounds.}
    \label{tab:task_scores}
    \setlength{\tabcolsep}{3pt}
    \renewcommand{\arraystretch}{1.05}
    \scriptsize
    \begin{tabular}{@{} l c *{21}{r} @{}}
        \toprule
        \textbf{Category} & \textbf{Task} & \rotatebox{70}{\textbf{Claude-Sonnet-3.7}} & \rotatebox{70}{\textbf{Claude-Sonnet-4}} & \rotatebox{70}{\textbf{Claude-Sonnet-4.5}} & \rotatebox{70}{\textbf{Llama-3.2-90b}} & \rotatebox{70}{\textbf{GPT-4o}} & \rotatebox{70}{\textbf{GPT-4.1}} & \rotatebox{70}{\textbf{GPT-5.2}} & \rotatebox{70}{\textbf{Doubao-seed-1.6}} & \rotatebox{70}{\textbf{Doubao-seed-1.8}} & \rotatebox{70}{\textbf{deepseek-v3.1}} & \rotatebox{70}{\textbf{deepseek-v3.2}} & \rotatebox{70}{\textbf{Kimi-K2}} & \rotatebox{70}{\textbf{Moonshot-v1}} & \rotatebox{70}{\textbf{Qwen-Flash}} & \rotatebox{70}{\textbf{Qwen-Turbo}} & \rotatebox{70}{\textbf{Qwen-Plus}} & \rotatebox{70}{\textbf{Qwen3-Max}} & \rotatebox{70}{\textbf{Qwen3-235B}} & \rotatebox{70}{\textbf{Qwen3-next-80B}} & \rotatebox{70}{\textbf{Gemini-2.5-flash}} & \rotatebox{70}{\textbf{Grok-3}} \\
        \midrule
        \multirow{3}{*}{\textbf{Atomic}} & Task 1 & \textbf{1.00} & \textbf{1.00} & \textbf{1.00} & \textbf{1.00} & \textbf{1.00} & \textbf{1.00} & \textbf{1.00} & \textbf{1.00} & \textbf{1.00} & 0.85 & 0.93 & \textbf{1.00} & \textbf{1.00} & 0.60 & \textbf{1.00} & 0.00 & \textbf{1.00} & 0.58 & 0.00 & \textbf{1.00} & \textbf{1.00} \\
         & Task 2 & \textbf{1.00} & \textbf{1.00} & \textbf{1.00} & \textbf{1.00} & \textbf{1.00} & \textbf{1.00} & \textbf{1.00} & \textbf{1.00} & \textbf{1.00} & 0.84 & 0.93 & \textbf{1.00} & \textbf{1.00} & \textbf{1.00} & \textbf{1.00} & \textbf{1.00} & \textbf{1.00} & \textbf{1.00} & \textbf{1.00} & \textbf{1.00} & \textbf{1.00} \\
         & Task 3 & \textbf{1.00} & \textbf{1.00} & \textbf{1.00} & \textbf{1.00} & \textbf{1.00} & \textbf{1.00} & \textbf{1.00} & \textbf{1.00} & \textbf{1.00} & \textbf{1.00} & \textbf{1.00} & \textbf{1.00} & \textbf{1.00} & \textbf{1.00} & \textbf{1.00} & \textbf{1.00} & \textbf{1.00} & \textbf{1.00} & \textbf{1.00} & \textbf{1.00} & \textbf{1.00} \\
        \midrule
        \multirow{10}{*}{\textbf{Sequential}} & Task 4 & 0.93 & \textbf{1.00} & \textbf{1.00} & \textbf{1.00} & \textbf{1.00} & \textbf{1.00} & \textbf{1.00} & \textbf{1.00} & \textbf{1.00} & \textbf{1.00} & \textbf{1.00} & 0.90 & 0.49 & 0.50 & \textbf{1.00} & \textbf{1.00} & \textbf{1.00} & \textbf{1.00} & \textbf{1.00} & \textbf{1.00} & \textbf{1.00} \\
         & Task 5 & 0.35 & 0.74 & 0.87 & 0.35 & 0.35 & 0.35 & 0.35 & 0.35 & 0.35 & 0.87 & 0.48 & 0.35 & 0.35 & 0.35 & 0.35 & 0.57 & 0.35 & \textbf{1.00} & 0.35 & 0.35 & 0.35 \\
         & Task 6 & \textbf{1.00} & 0.92 & 0.60 & \textbf{1.00} & 0.95 & \textbf{1.00} & \textbf{1.00} & \textbf{1.00} & \textbf{1.00} & \textbf{1.00} & \textbf{1.00} & \textbf{1.00} & \textbf{1.00} & \textbf{1.00} & 0.50 & \textbf{1.00} & \textbf{1.00} & \textbf{1.00} & \textbf{1.00} & \textbf{1.00} & \textbf{1.00} \\
         & Task 7 & 0.76 & 0.93 & \textbf{0.99} & 0.86 & 0.94 & 0.96 & \textbf{0.99} & 0.93 & 0.77 & 0.24 & 0.22 & 0.93 & 0.43 & 0.07 & 0.30 & 0.30 & 0.77 & 0.07 & 0.93 & 0.94 & 0.93 \\
         & Task 8 & \textbf{1.00} & 0.76 & 0.74 & 0.73 & 0.42 & 0.42 & 0.56 & 0.39 & 0.39 & 0.50 & 0.38 & 0.95 & 0.41 & 0.44 & 0.37 & 0.96 & 0.38 & 0.88 & 0.33 & 0.95 & 0.99 \\
         & Task 9 & 0.22 & 0.37 & 0.25 & 0.29 & 0.43 & 0.33 & 0.33 & 0.21 & 0.23 & 0.21 & \textbf{0.46} & 0.11 & 0.28 & 0.00 & 0.00 & 0.04 & 0.20 & 0.11 & 0.06 & 0.35 & 0.20 \\
         & Task 10 & 0.24 & 0.35 & \textbf{0.74} & 0.28 & 0.35 & 0.35 & 0.35 & 0.35 & 0.35 & 0.35 & 0.28 & 0.20 & 0.35 & 0.35 & 0.35 & 0.17 & 0.35 & 0.35 & 0.35 & 0.35 & 0.17 \\
         & Task 11 & \textbf{1.00} & \textbf{1.00} & \textbf{1.00} & \textbf{1.00} & \textbf{1.00} & \textbf{1.00} & \textbf{1.00} & \textbf{1.00} & \textbf{1.00} & \textbf{1.00} & \textbf{1.00} & \textbf{1.00} & 0.90 & \textbf{1.00} & \textbf{1.00} & \textbf{1.00} & \textbf{1.00} & \textbf{1.00} & \textbf{1.00} & \textbf{1.00} & \textbf{1.00} \\
         & Task 12 & 0.35 & \textbf{1.00} & 0.35 & 0.23 & 0.35 & 0.35 & 0.35 & 0.35 & 0.35 & 0.33 & 0.25 & 0.32 & 0.23 & 0.17 & 0.35 & 0.17 & 0.35 & 0.23 & 0.35 & 0.35 & 0.35 \\
         & Task 13 & \textbf{1.00} & \textbf{1.00} & \textbf{1.00} & \textbf{1.00} & \textbf{1.00} & \textbf{1.00} & \textbf{1.00} & \textbf{1.00} & \textbf{1.00} & \textbf{1.00} & \textbf{1.00} & \textbf{1.00} & 0.89 & \textbf{1.00} & \textbf{1.00} & \textbf{1.00} & \textbf{1.00} & \textbf{1.00} & \textbf{1.00} & \textbf{1.00} & \textbf{1.00} \\
        \midrule
        \multirow{4}{*}{\textbf{Parallel}} & Task 14 & 0.90 & \textbf{1.00} & \textbf{1.00} & \textbf{1.00} & \textbf{1.00} & 0.90 & \textbf{1.00} & \textbf{1.00} & \textbf{1.00} & \textbf{1.00} & \textbf{1.00} & \textbf{1.00} & \textbf{1.00} & 0.75 & 0.80 & 0.75 & 0.95 & \textbf{1.00} & 0.90 & \textbf{1.00} & \textbf{1.00} \\
         & Task 15 & \textbf{1.00} & \textbf{1.00} & \textbf{1.00} & 0.52 & \textbf{1.00} & \textbf{1.00} & \textbf{1.00} & \textbf{1.00} & \textbf{1.00} & 0.87 & \textbf{1.00} & \textbf{1.00} & 0.75 & 0.69 & \textbf{1.00} & 0.65 & \textbf{1.00} & 0.82 & 0.84 & \textbf{1.00} & 0.97 \\
         & Task 16 & 0.50 & 0.50 & 0.50 & \textbf{1.00} & 0.50 & 0.50 & 0.50 & 0.50 & 0.50 & 0.50 & 0.50 & 0.60 & \textbf{1.00} & \textbf{1.00} & 0.75 & \textbf{1.00} & 0.50 & \textbf{1.00} & \textbf{1.00} & 0.50 & 0.60 \\
         & Task 17 & \textbf{1.00} & 0.70 & 0.45 & 0.55 & 0.85 & 0.90 & 0.95 & \textbf{1.00} & 0.87 & 0.36 & 0.33 & 0.76 & 0.33 & \textbf{1.00} & \textbf{1.00} & 0.79 & 0.25 & 0.33 & 0.60 & 0.42 & \textbf{1.00} \\
        \midrule
        \multirow{5}{*}{\textbf{S-P Comp.}} & Task 18 & 0.67 & 0.73 & 0.73 & 0.80 & 0.80 & 0.80 & 0.87 & 0.93 & \textbf{1.00} & 0.93 & 0.93 & 0.93 & 0.80 & 0.97 & 0.68 & \textbf{1.00} & \textbf{1.00} & 0.87 & \textbf{1.00} & 0.67 & \textbf{1.00} \\
         & Task 19 & 0.28 & \textbf{1.00} & 0.66 & 0.28 & 0.31 & 0.31 & 0.29 & 0.28 & 0.41 & 0.28 & 0.28 & 0.83 & 0.43 & 0.45 & 0.28 & 0.28 & 0.28 & 0.28 & 0.37 & 0.82 & 0.76 \\
         & Task 20 & \textbf{1.00} & 0.93 & 0.93 & 0.56 & \textbf{1.00} & \textbf{1.00} & 0.97 & \textbf{1.00} & 0.93 & 0.95 & 0.84 & 0.94 & 0.88 & 0.88 & 0.53 & \textbf{1.00} & 0.91 & \textbf{1.00} & \textbf{1.00} & 0.95 & 0.84 \\
         & Task 21 & \textbf{1.00} & 0.79 & 0.28 & 0.75 & 0.53 & 0.60 & 0.53 & \textbf{1.00} & 0.49 & 0.70 & 0.89 & 0.37 & 0.69 & 0.56 & 0.75 & 0.88 & 0.56 & 0.65 & 0.41 & 0.62 & 0.65 \\
         & Task 22 & 0.50 & 0.50 & 0.50 & 0.58 & 0.51 & 0.50 & 0.50 & 0.61 & 0.50 & 0.54 & 0.56 & 0.52 & 0.64 & 0.60 & 0.61 & 0.62 & 0.50 & \textbf{0.75} & 0.50 & 0.54 & 0.64 \\
        \midrule
        \multirow{8}{*}{\shortstack{\textbf{Lifecycle-}\\\textbf{Scoped T.}}} & Task 23 & 0.40 & 0.80 & 0.00 & 0.75 & 0.90 & \textbf{1.00} & 0.80 & 0.00 & 0.00 & 0.00 & 0.00 & 0.38 & 0.38 & 0.00 & 0.00 & 0.81 & 0.38 & 0.31 & 0.00 & 0.80 & \textbf{1.00} \\
         & Task 24 & \textbf{1.00} & \textbf{1.00} & 0.45 & 0.86 & 0.81 & 0.92 & \textbf{1.00} & 0.33 & 0.36 & 0.47 & 0.56 & \textbf{1.00} & 0.59 & 0.51 & 0.57 & 0.67 & \textbf{1.00} & 0.45 & 0.48 & 0.90 & \textbf{1.00} \\
         & Task 25 & \textbf{1.00} & \textbf{1.00} & \textbf{1.00} & 0.77 & 0.93 & \textbf{1.00} & \textbf{1.00} & \textbf{1.00} & 0.68 & 0.82 & 0.87 & 0.51 & 0.36 & 0.38 & 0.54 & \textbf{1.00} & \textbf{1.00} & 0.84 & 0.50 & \textbf{1.00} & \textbf{1.00} \\
         & Task 26 & 0.70 & 0.41 & 0.44 & 0.43 & 0.43 & 0.46 & 0.45 & 0.70 & \textbf{0.75} & 0.43 & 0.42 & 0.59 & 0.46 & 0.45 & 0.45 & 0.43 & 0.43 & 0.45 & 0.42 & 0.57 & 0.50 \\
         & Task 27 & 0.82 & 0.92 & \textbf{0.95} & 0.50 & 0.83 & 0.75 & 0.70 & 0.72 & 0.72 & 0.66 & 0.49 & 0.50 & 0.39 & 0.50 & 0.33 & 0.58 & 0.92 & 0.84 & 0.50 & 0.92 & 0.67 \\
         & Task 28 & 0.83 & \textbf{1.00} & 0.71 & 0.58 & 0.51 & 0.47 & 0.47 & 0.50 & 0.47 & 0.79 & 0.49 & 0.58 & 0.64 & 0.50 & 0.76 & 0.50 & 0.92 & 0.82 & 0.50 & 0.86 & 0.47 \\
         & Task 29 & \textbf{1.00} & 0.75 & 0.94 & 0.48 & 0.65 & 0.66 & 0.58 & 0.81 & 0.39 & 0.80 & 0.39 & 0.39 & 0.46 & 0.39 & 0.38 & 0.39 & 0.92 & 0.64 & 0.94 & 0.66 & \textbf{1.00} \\
         & Task 30 & 0.70 & 0.74 & 0.75 & 0.47 & 0.52 & 0.60 & 0.61 & 0.39 & 0.39 & 0.83 & 0.87 & 0.43 & 0.45 & 0.38 & 0.34 & 0.39 & 0.79 & 0.82 & \textbf{0.90} & 0.74 & 0.87 \\
        \midrule
        \midrule
        \multicolumn{2}{l}{\textbf{Micro-Avg}} & \shortstack{0.77\\$\pm$0.03} & \textbf{\shortstack{0.83\\$\pm$0.02}} & \shortstack{0.73\\$\pm$0.03} & \shortstack{0.69\\$\pm$0.04} & \shortstack{0.73\\$\pm$0.02} & \shortstack{0.74\\$\pm$0.01} & \shortstack{0.74\\$\pm$0.02} & \shortstack{0.71\\$\pm$0.01} & \shortstack{0.66\\$\pm$0.03} & \shortstack{0.67\\$\pm$0.03} & \shortstack{0.65\\$\pm$0.01} & \shortstack{0.70\\$\pm$0.02} & \shortstack{0.62\\$\pm$0.01} & \shortstack{0.58\\$\pm$0.03} & \shortstack{0.60\\$\pm$0.01} & \shortstack{0.66\\$\pm$0.01} & \shortstack{0.72\\$\pm$0.01} & \shortstack{0.70\\$\pm$0.02} & \shortstack{0.64\\$\pm$0.01} & \shortstack{0.78\\$\pm$0.01} & \shortstack{0.80\\$\pm$0.03} \\
        \multicolumn{2}{l}{\textbf{Macro-Avg}} & \shortstack{0.81\\$\pm$0.03} & \textbf{\shortstack{0.84\\$\pm$0.01}} & \shortstack{0.75\\$\pm$0.03} & \shortstack{0.73\\$\pm$0.04} & \shortstack{0.77\\$\pm$0.02} & \shortstack{0.78\\$\pm$0.01} & \shortstack{0.78\\$\pm$0.02} & \shortstack{0.77\\$\pm$0.01} & \shortstack{0.72\\$\pm$0.02} & \shortstack{0.70\\$\pm$0.03} & \shortstack{0.70\\$\pm$0.01} & \shortstack{0.76\\$\pm$0.03} & \shortstack{0.69\\$\pm$0.01} & \shortstack{0.66\\$\pm$0.05} & \shortstack{0.68\\$\pm$0.01} & \shortstack{0.69\\$\pm$0.01} & \shortstack{0.75\\$\pm$0.01} & \shortstack{0.73\\$\pm$0.02} & \shortstack{0.67\\$\pm$0.02} & \shortstack{0.80\\$\pm$0.01} & \shortstack{0.84\\$\pm$0.03} \\
        \bottomrule
    \end{tabular}
\end{sidewaystable*}

  \section{Limitations and Future Work}\label{appendix:limitations}

\textbf{Context Length Constraints:}
Our approach is constrained by the finite context window of contemporary LLMs, which places an upper bound on total prompt and output sequence length.
As embodied systems evolve and the function interface library expands, prompts may surpass this capacity, potentially causing the omission of earlier or less relevant function definitions.
This limitation can be readily addressed through Retrieval-Augmented Generation~(RAG)~\cite{fan2024survey}, which enables dynamic retrieval of relevant function interfaces from an external knowledge base for selective incorporation into prompts.

\textbf{Non-Full-Duplex Interaction:}
Current LLM inference mechanisms do not support full-duplex communication during generation, which prevents real-time incorporation of environmental state updates or function execution feedback within a single conversational turn.
Our approach employs an interruption-based mechanism that terminates ongoing generation to handle state updates.
Although OpenAI's Realtime API~\cite{openai2024realtime} introduces bidirectional audio streaming, its speech-to-speech protocol remains incompatible with our text-based XML \textit{function token} schema.

\textbf{Specialized Embodied Programming Models:}
Future work should investigate fine-tuning LLMs and LMMs, or training smaller VLM and VLA models specifically for mixed \textit{function token} and natural language generation.
Training or fine-tuning such models would instill inductive biases for both proficient robotic function interface utilization and enhanced real-world understanding, thereby improving their ability to generate accurate XML \textit{function tokens} and robust multi-component behavior orchestration.
Moreover, edge deployment of these models could reduce inference latency and improve real-time responsiveness, enhancing the autonomy and reliability of robotic systems.
This in turn supports on-device streaming embodied programming without compromising language understanding or behavioral control.

\textbf{Security and Robustness Enhancement:}
Security mechanisms become essential as \algname integrates LLMs with physical robotic platforms.
Recent research has identified attack vectors including prompt manipulation~\cite{liu2024exploring} and contextual backdoor attacks targeting code generation~\cite{liu2025compromising} in LLM-driven embodied systems.
Future work should establish defense mechanisms against prompt injection attacks on XML \textit{function token} generation and contextual poisoning of interface code, enforce safety constraints within the multi-channel scheduler, and enable real-time monitoring for hazardous behaviors.
\fi

\end{document}